%% file: main.tex
\documentclass[10pt,twocolumn,letterpaper]{article}

\usepackage[pagenumbers]{wacv} % To force page numbers, e.g. for an arXiv version
\input{preamble}
\input{math_commands.tex}

\definecolor{wacvblue}{rgb}{0.21,0.49,0.74}
\usepackage[pagebackref,breaklinks,colorlinks,allcolors=wacvblue]{hyperref}

\def\wacvPaperID{2464} % *** Enter the WACV Paper ID here
\def\confName{WACV}
\def\confYear{2026}

\title{DreamMakeup: Face Makeup Customization using Latent Diffusion Models}

\author{
Geon Yeong Park*\textsuperscript{1} \qquad Inhwa Han*\textsuperscript{1} \qquad Serin Yang*\textsuperscript{1} \qquad Yeobin Hong*\textsuperscript{1} \qquad Seongmin Jeong\textsuperscript{2}\\
Heechan Jeon\textsuperscript{2} \qquad Myeongjin Goh\textsuperscript{2} \qquad Sung Won Yi\textsuperscript{2} \qquad Jin Nam\textsuperscript{2} \qquad Jong Chul Ye\textsuperscript{1}\\
\textsuperscript{1}KAIST \qquad \textsuperscript{2}Amorepacific\\
\textsuperscript{*}Equal contribution
}

\begin{document}

\twocolumn[{%
\renewcommand\twocolumn[1][]{#1}%
\maketitle
\vspace{-0.9cm}
\begin{center}
    \centering
    \captionsetup{type=figure}
    \includegraphics[width=\textwidth]{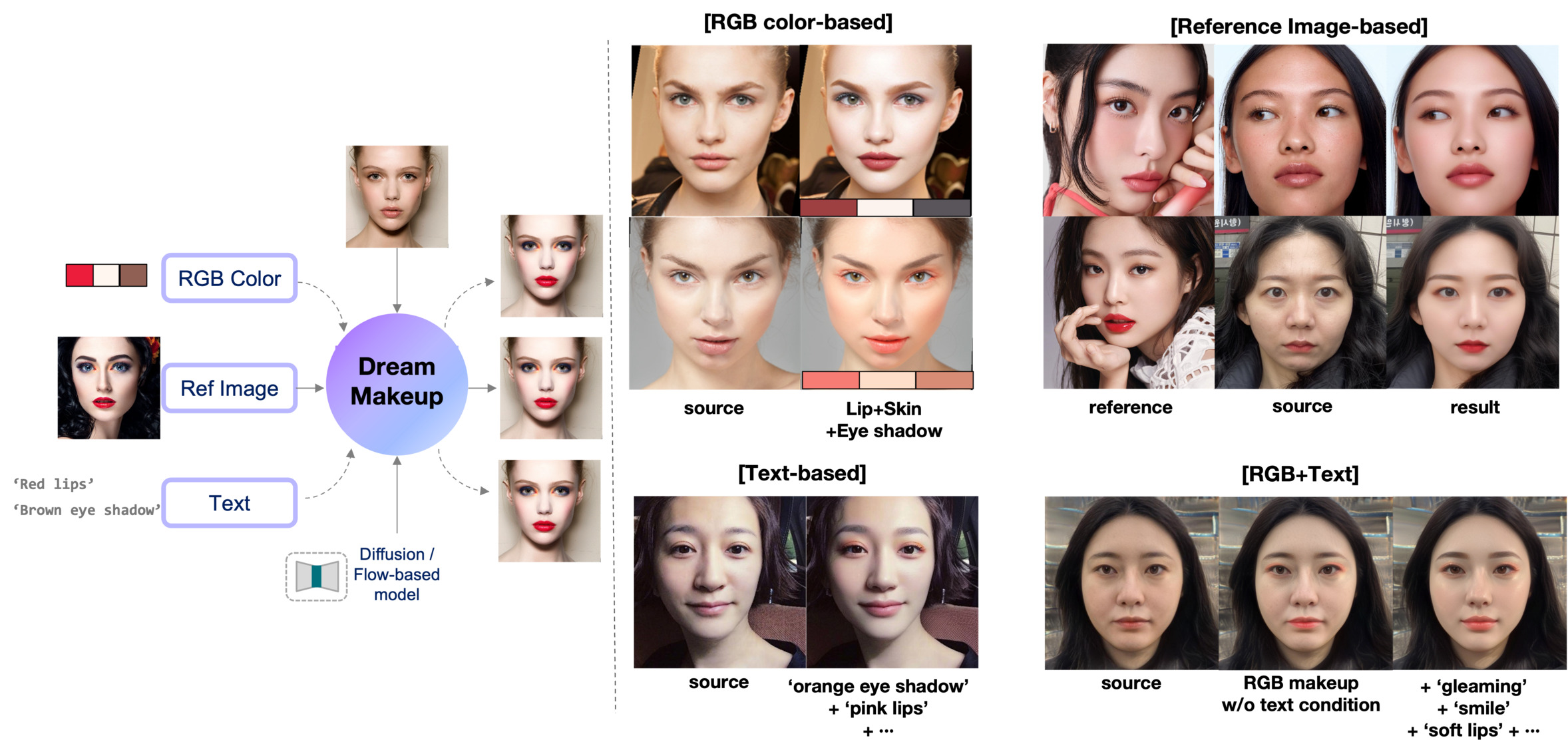}
    \captionof{figure}{
        We present \textbf{DreamMakeup}, a training-free diffusion framework that generates high-fidelity makeup results by integrating diverse user inputs such as RGB colors, reference images, and text prompts. Our method produces high-quality, customized makeup while preserving facial identity, without requiring any fine-tuning. Please zoom in for detailed inspection.}
    \vspace{-0.10cm}
    \label{fig: teaser}
\end{center}%
}]

\begin{abstract}
The exponential growth of the global makeup market has paralleled advancements in virtual makeup simulation technology. Despite the progress led by GANs, their application still encounters significant challenges, including training instability and limited customization capabilities. Addressing these challenges, we introduce {\em DreamMakup} -- a novel training-free Diffusion model based Makeup Customization method, leveraging the inherent advantages of diffusion models for superior controllability and precise real-image editing. DreamMakeup employs early-stopped DDIM inversion to preserve the facial structure and identity while enabling extensive customization through various conditioning inputs such as reference images, specific RGB colors, and textual descriptions. Our model demonstrates notable improvements over existing GAN-based and recent diffusion-based frameworks -- improved customization, color-matching capabilities, identity preservation and compatibility with textual descriptions or LLMs with affordable computational costs.
%Project page is available at \href{https://dreammakeup.github.io/}{here}.
\vspace{-0.2cm}
\end{abstract}

\begin{figure*}[htbp]
    \centering
    \includegraphics[width=\textwidth]{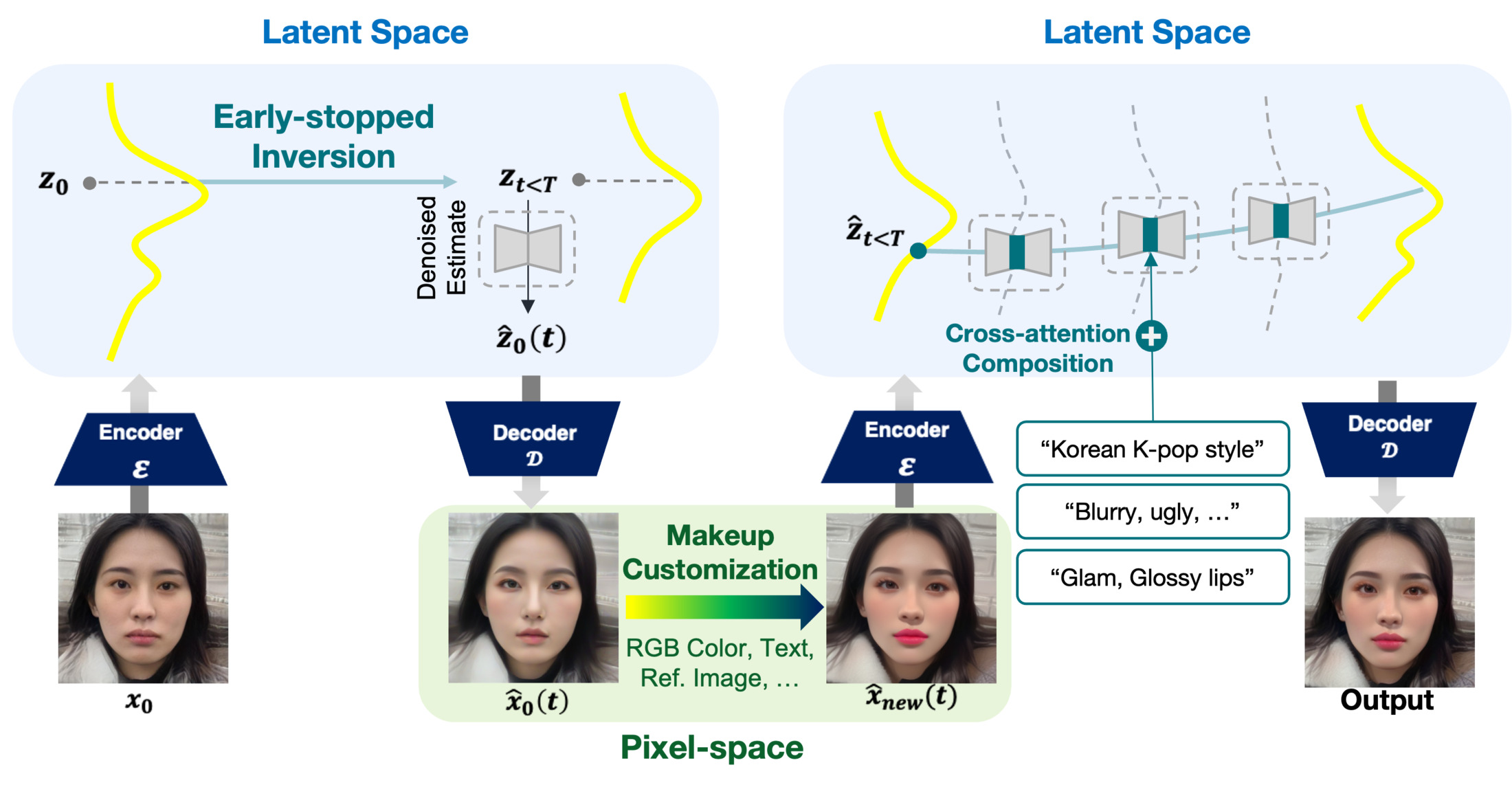}
    \caption{
    \textbf{Overview of \textit{DreamMakeup} pipeline}. The key principle of our framework is to apply fine-grained guidance in high-dimensional pixel-domain during reverse sampling. After local makeup customization in pixel space, text prompts are leveraged to harmonize such local variations with a consistent global style in latent cross-attention space.
    }
    \vspace{-0.5cm}
    \label{fig: overall}
\end{figure*}

\section{Introduction}
\label{sec:intro}
The global makeup market size is valued at billions of dollars, and virtual makeup simulation technology is considered to be a rapidly growing sector within the beauty industry. Besides its industrial importance, face makeup customization is also an interesting problem in terms of generative modeling and editing. Specifically, one may have to disentangle and stylize each facial attirbute in their independent style, while its composition should be well harmonized. 

So far, virtual face makeup modeling is mainly driven by generative adversarial networks (GANs) \cite{li2018beautygan, jiang2019psgan, deng2021spatially, liu2023gan, yang2022elegant}. Despite its advancements, GAN-based frameworks face inherent instability in adversarial training, leading to several limitations. Furthermore, existing GAN-based methods are not fully customizable and lack controllability. Specifically, most of these frameworks \textit{only} support makeup transfer tasks, inherently requiring reference target images. In many business contexts, users may seek to simulate facial makeup with a more degree of freedom, e.g. test with specific RGB colors of new cosmetic products, or linguistic descriptions such as \texttt{``Glam makeup style"}, etc. 

In response to these challenges, this paper explores the adoption of diffusion models, recognized for their superior controllability and real-image editing capabilities. Diffusion models offer several advantages for facial beauty simulation. For instance, we can control the reverse sampling process using the enriched text-conditions. Moreover, it supports various style customization using LoRAs \cite{hu2021lora} supported by the vibrant user community. By employing techniques such as DDIM inversion \cite{song2020denoising, mokady2023null}, diffusion models well preserve the overall structure and subject identity of given facial images, while retaining rich editing capabilities.

However, straightforward application of generic diffusion-based editing algorithms is inadequate for customized makeup, often failing to perform accurate color-matching and natural makeup in specified facial regions (\cref{fig:diffusion_editing}). Consequently, while several specialized diffusion-based makeup frameworks have been introduced \cite{zhang2024stable, sun2024diffam, jin2025toward, lu2024makeupdiffuse}, many of them still face similar constraints with GANs; many remain limited to reference-based makeup transfer, failing to leverage more diverse conditioning inputs. Additionally, they require expensive training/fine-tuning, potentially restricting the generative capacity of powerful prior models, and often lose structural identity (Stable-Makeup in \cref{fig:Reference_comparison}).

%To mitigate these problems, we leverage diffusion models, lauded for their exceptional controllability and sampling guidance capabilities in real-image editing tasks \cite{hertz2022prompt, tumanyan2023plug, park2024energy}. There are several advantages to using diffusion models in facial makeup and attribute manipulation: First, there are several off-the-shelf foundational diffusion models like stable diffusion which have a rich generative prior. There exists a big user society that customizes the pre-trained diffusion models in their way and style, mainly empowered by LoRA \cite{hu2021lora}, providing a rich foundation for makeup customization. Moreover, one may invert the given image to an initial noise vector by using DDIM inversion \cite{song2020denoising, mokady2023null} which is an iterative simulation of the forward ODE flow. From these inverted codes, one can retain the rich editing capabilities, e.g. makeup simulation, while preserving the overall structure and subject identity of the input face image. In diffusion models, such an editing process can be guided by several modalities, such as text prompts, semantic maps, etc. In virtual makeup simulation, we may utilize various conditioning to guide facial makeup. 

To this end, we introduce DreamMakeup, a novel \textit{training-free} diffusion-based makeup customization distinguished by its advanced customization and superior identity preservation. 
Dreammakeup is fully compatible with a variety of conditionings to steer the makeup process, ranging from reference images and specific RGB colors to textual descriptions of desired makeup looks. 
As shown in \cref{fig: overall}, given pre-trained latent diffusion models (LDM), we commence by inverting facial images $\xb_0$ into latents $\zb_t$ through early-stopped DDIM inversion. Subsequently, we approximate the denoised estimate $\hat\zb_0(t):=\mathbb{E}[\zb_0 | \zb_t]$ and decode it back into the pixel space, preserving the facial structure attributed to the inversion process. Then, we stylize these facial representations in pixel space through transformations such as histogram matching, RGB color matching, or warping, toward a targeted makeup style. Finally, resuming the sampling process from these transformed representations, with advanced cross-attention control and interpolation-guided sampling, yields harmonized and coherent makeup outcomes.
Our contributions are summarized as follows:

%Specifically, global makeup stylization and refinement is conducted through reverse sampling with EBCQ \cite{park2024energy}. 
\begin{itemize}[noitemsep]
    \item We introduce \textit{DreamMakeup}, a novel diffusion-based human face makeup framework that caters to a wide range of user preferences including text descriptions, colors, and reference images.
    %To our knowledge, we mark a pioneering case of leveraging large-scale foundational diffusion prior and its LoRA variations in face makeup customization.
    \item DreamMakeup is computationally affordable as it does not fundamentally require fine-tuning. Moreover, we early-stop DDIM inversion process to preserve the facial structures which further accelerates inference ($< 4$ seconds for color transfer w/ SD v1.5 \citep{rombach2022high}, GeForce RTX 4090). 
    \item DreamMakeup outperforms real-world global AI makeup services in color maekup task, and state-of-the-art Diffusion/GAN-based frameworks in makeup transfer tasks. Furthermore, we demonstrate that our framework can be easily integrated with other foundational models, including Large Language Models (LLMs), facial classifiers, LoRAs or even other makeup diffusion models.
\end{itemize}

% EBCA

%then sampling with ebcq. this blends attention maps which is compatible with various prompts. 

%Various applications: reference RGB color, image, LLM. Also reference description. 

\section{Related Works}
%\subsection{Makeup transfer}
Makeup transfer aims to modify a facial image to reflect a chosen makeup style, with numerous approaches developed using Generative Adversarial Networks (GANs). BeautyGAN \cite{li2018beautygan} employed histogram matching to preserve color from the reference image. LADN \cite{gu2019ladn} used local discriminators for heavy makeup. PSGAN \cite{jiang2019psgan} enhanced style controllability through matrices and addressed pose misalignments with attention mechanisms. SCGAN \cite{deng2021spatially} tackled pose issues with style codes. RamGAN \cite{xiang2022ramgan} improved makeup transfer with regional attention; EleGANt \cite{yang2022elegant} controlled arbitrary regions using attention mechanisms, reducing computations.

However, GAN-based methods typically require large datasets of makeup and no-makeup images for training and reference images for inference, limiting application diversity and customizability. This is in line with recent diffusion-based methods \cite{lu2024makeupdiffuse, zhang2024stable, jin2025toward} which often focus on adapting diffusion models for reference-based makeup with fine-tuning. Moreover, due to the stochastic generative nature of diffusion models, these works often fail to preserve the structural identity of input images (e.g., Stable Makeup in \cref{fig:Reference_comparison}). 

DreamMakeup overcomes these limitations by utilizing the powerful prior of foundational diffusion models, DDIM inversion and open-source LoRAs for image generation, enabling replication of makeup styles from references, manipulation via RGB or text prompts. This approach enhances controllability without relying on large datasets or fine-tuning.

%For more discussions on related works, please refer to the appendix
% local : pixel domain, global: text guidance sampling
% pixel domain에서 guidance를 fine-grained guidance를 줄 수 있음. 이는 latent에서 guidance를 주는 것이 아니기 때문에 차별점이 있음.
% diffusion이 edge같은 부분을 noise를 더하면서 discontinuity를 잘 해결해 줌
% mask를 soft하게 만들어서 discontinuity를 해결함
% constraint가 경계선이 딱 나뉘는데, 이게 reverse sampling 과정에서 자연스럽게 해결됨.
% generation image처럼 안보인다는게 다른 부분은 다 유지가 됨. 이런 것들이 key technology. 
% Glossy lips가 유지가 되는 것들은, diffusion model 기술을 통해서 자연스럽게 합치는 것
% 확대 inset 자연스러움. 
% GAN: 얼굴이 커질 수도 있음. 어색한 부분이 있을 수 있음. 자기 사진을 넣어주고 reference image를 넣어줬을 때 해당 스타일로 transform이 될 수 있음. 처음 결과와 나중 결과를 뺐을 때 그 차이 map을 visualize하면서 plot할 수 있음. GAN을 이용해서 생성했을 때 품질을 평가하는 지표들이 있을 수 있음. GAN type의 모델들은 pose variation에 취약함. 그러나 그런 outlier는 그대로 유지하면서 transformation이 가능할 것.
% 원본 아이덴티티를 유지하면서 화장을 하는게 가장 좋은 장점.
% 원본 사이즈의 비례가 딱 맞음. GAN은 좌우 밸런스가 잘 안 맞음. 수치적으로 그런 것들.
% 

% 기술특허와는 다르게 모든 기술을 전부 공개하는건 어려움
% 소프트웨어 특허는 특히 공개가 어려움
% analogous to AdaIN.
% 전체 pipeline 자체가 business의 내용이 됨.
% 특허는 기술 부분에서 가출원이라도 하도록

% 피드백: 팔자주름, 눈썹, 입술 등 구조 유지 할 수 있도록
% 5월부터 후속연구 논의/

\section{Preliminary}
Diffusion models aim to generate samples from the Gaussian noise through iterative denoising processes. Since pixel-space diffusion models are computationally heavy, the latent diffusion model (LDM) \cite{rombach2022high} operates the diffusion process on latent space instead of pixel space. 
Given a pixel-space clean sample $\xb \sim p_{\data}(\xb)$, \cite{rombach2022high} leverages an autoencoder
\begin{equation}
    \label{eq: autoencoder}
        \Ec: \mathbb{R}^d \rightarrow \mathbb{R}^k, \Dc: \mathbb{R}^k \rightarrow \mathbb{R}^d, \xb \simeq \Dc(\Ec(\xb)), \forall \xb \sim p_{\text{data}}(\xb), 
\end{equation}
where $\Ec$ is the encoder, $\Dc$ is the decoder, and dimension of the latent space $k < d$. After training $\Ec$, $\Dc$, forward and reverse diffusion process can be defined within the latent space $\zb = \Ec(\xb)$.

The forward process is defined as a Markov chain, characterized by forward conditional densities:
\begin{align}
\label{eq: diffusion forward density}
p(\xb_t | \xb_{t-1}) &= \Nc (\xb_t | \beta_t \xb_{t-1}, (1-\beta_t) I), \\ 
p_t(\xb_t | \xb_0) &= \Nc (\xb_t | \sqrt{\alphabar} \xb_0, (1-\alphabar) I),
\end{align}
with $\zb_{t} \in \mathbb{R}^k$ representing the noisy latent variable at a timestep $t \leq T$ that has the same dimension as $\zb_0 = \Ec(\xb_0)$ for $\xb_0 \sim p_{\text{data}}(\xb)$, and $\beta_t$ denotes an increasing sequence of noise schedule where $\alpha_t := 1 - \beta_t$ and $\alphabar_t := \Pi_{i=1}^t \alpha_i$. The goal of training LDM is to obtain a residual denoiser $\epsilonb_{\theta^*}$: 
\begin{equation}
    \label{eq: epsilon matching}
    \theta^*= \arg\min_{\theta} \Eb_{\Ec(\xb_0), t, \epsilonb \sim \Nc(0, I)} \big[ \norm{\epsilonb_{\theta} (\zb_t, t) - \epsilonb} \big].
\end{equation}
%where $\epsilonb_{\theta^*}(\xb_t, t) \simeq \epsilonb = \frac{\xb_t - \sqrt{\alphabar_t} \xb_0 }{\sqrt{1-\alphabar}}$. 

\noindent The reverse sampling from $q(\zb_{t-1}|\zb_t, \epsilonb_{\theta^*}(\zb_t, t))$ is then
%  For the reverse process, with the learned noise prediction network $\boldsymbol{\epsilon}_\theta$, we can predict the noisy sample of previous timestep $\x_{t-1}$:   
\begin{align}
    \label{eq: reverse sampling}
    \zb_{t-1}=\frac{1}{\sqrt{\alpha_t}}\Big{(}\zb_t-\frac{1-\alpha_t}{\sqrt{1-\bar{\alpha}_t}}\boldsymbol{\epsilon}_{\theta^*}(\zb_t, t)\Big{)}+\tilde{\beta}_t \epsilonb,
\end{align}
where $\epsilonb \sim\Nc(0,\Ib)$ and $\tilde{\beta}_t := \frac{1 - \alphabar_{t-1}}{1 - \alphabar_t} \beta_t$. For simplicity, we will omit $*$ in $\theta^*$.
After reverse sampling, the generated latent $\tilde{\zb}_0$ is decoded to the pixel space as $\tilde{\xb}_0 = \Dc(\tilde{\zb}_0)$.

To accelerate sampling, DDIM \cite{song2020denoising} proposes an alternative sampling method:
\begin{equation}
    \label{eq: ddim sampling}
    \zb_{t-1} = \sqrt{\alphabar_{t-1}} \hat{\zb}_0(t) + \sqrt{ 1 - \alphabar_{t-1} - \eta^2 \tilde{\beta_t}^2} \epsilonb_{\theta} (\zb_t, t) + \eta \tilde{\beta}_t \epsilonb,
\end{equation}
where $\eta \in [0,1]$ is a stochasticity parameter, and $\hat{\zb}_0(t)$ is the denoised estimate which can be equivalently derived using Tweedie's formula \cite{efron2011tweedie}:
\begin{equation}
    \label{eq: Tweedie}
    \hat{\zb}_0(t) := \frac{1}{\sqrt{\alphabar_t}} (\zb_t - \sqrt{1-\alphabar_t} \epsilonb_{\theta} (\zb_t, t)).
\end{equation}
For text-guided sampling, we train the diffusion model with textual embedding $c$. We will often omit $c$ from $\epsilonb_{\theta} (\xb_t, t, c)$ to avoid notational complexity.

\section{DreamMakeup}%Diffusion-based Makeup Customization}

Our \textit{primary} focus is on daily, realistic and aesthetic makeup (\cref{fig: teaser}a), rather than complex and excessive makeup styles/transfer.
Specifically, given an input non-makeup image $\xb_0$, our main goal is to (\textbf{a}) customize the makeup style with coarse (e.g. RGB color) to fine (e.g. reference makeup image) level information, while (\textbf{b}) preserving the overall facial structure and subject identity to the greatest extent. Furthermore, we aim for applications beyond its primary scope. It can be extended to extreme makeup styles (\cref{fig:extreme_makeup}), and its capacity to handle customized user inputs enables diverse applications such as hair dyeing (\cref{fig:RGB_application}).
%Moreover, %as shown in Appendix with experimental details and pseudo-code,
Finally,  we are interested in integrating with LLMs or facial classifiers, paving a new path for virtual makeup pipeline design. 
%Related experimental details and pseudo-code are available in the Appendix.

To achieve this, one of our primary contributions is to integrate a pixel-space makeup customization during reverse sampling process given the decoded intermediate estimates $\hat{\xb}_0(t) =  \Dc(\hat{\zb}_0(t))$. 
Furthermore, \textit{DreamMakeup} leverages various user preferences for conditional guidance, e.g. target color, reference image, and textual make-up description. As shown in \cref{fig: overall}, the customization process consists of three main phases: (\textbf{1}) early-stopped DDIM inversion to impose structural consistency, (\textbf{2}) pixel-space customization to guide the sampling process towards target makeup style, and (\textbf{3}) reverse sampling with cross attention composition and interpolation guidance to accommodate complex textual makeup descriptions simultaneously. 
% \subsection{LLM and Facial Classifier Integration}

% More details are as follows.

\begin{figure}[t!]
    \centering
    \includegraphics[width=\linewidth]{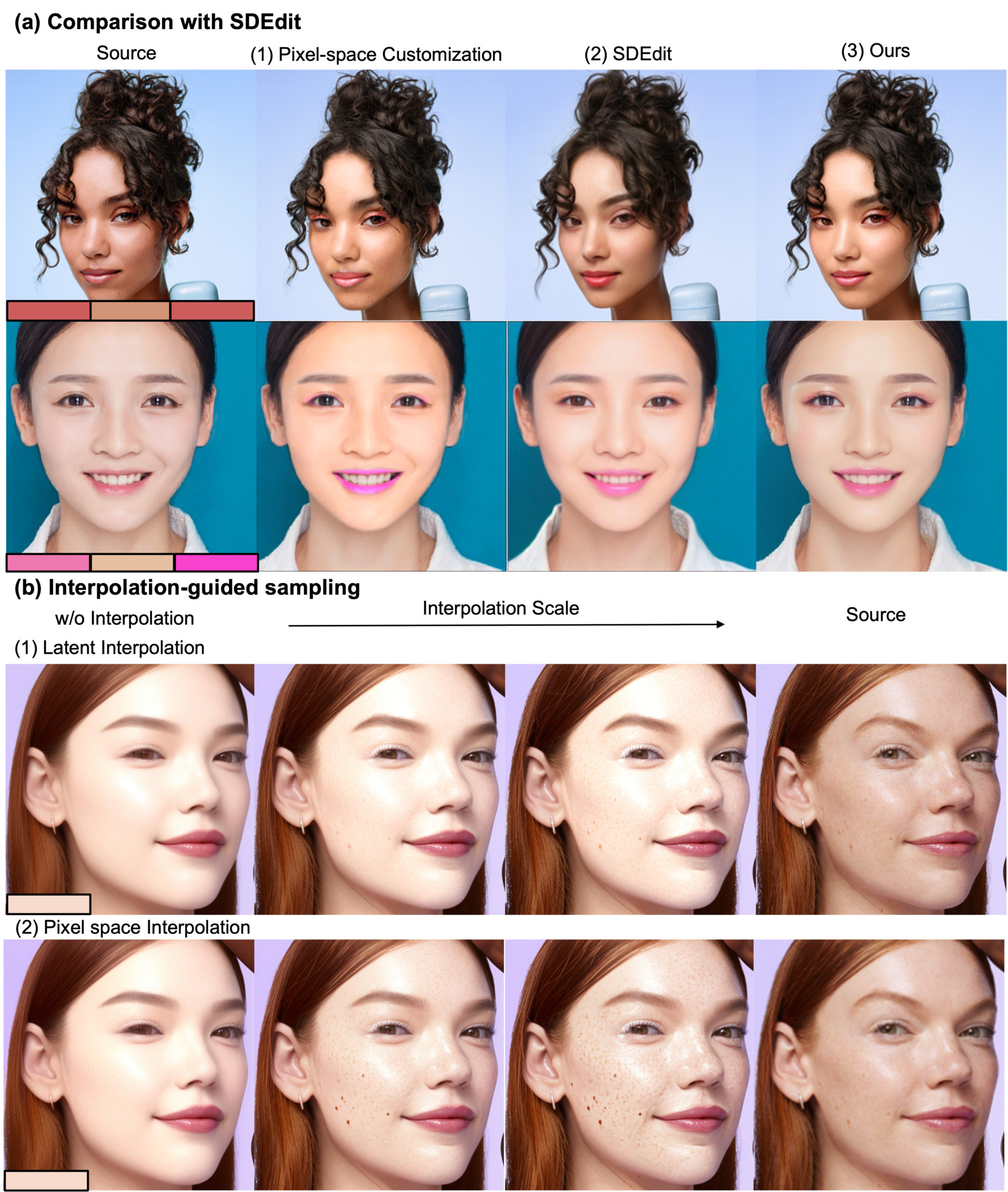}
    \caption{(\textbf{a}) Direct pixel-space customization results in color inconsistencies, while SDEdit (strength=0.2) degrades the subject's identity. In contrast, our method applies the makeup faithfully while preserving identity. (\textbf{b}) The impact of the interpolation domain during reverse sampling. Latent-space interpolation (\textbf{b-1}) effectively preserves fine-grained facial details, whereas pixel-space interpolation (\textbf{b-2}) introduces significant visual artifacts. Further details are provided in \cref{sec: EBCQ}.}
    \vspace{-0.4cm}
    \label{fig:sdedit}
\end{figure}

% prompt ablation results
\begin{figure*}[t!]
    \centering
    \includegraphics[width=\textwidth]{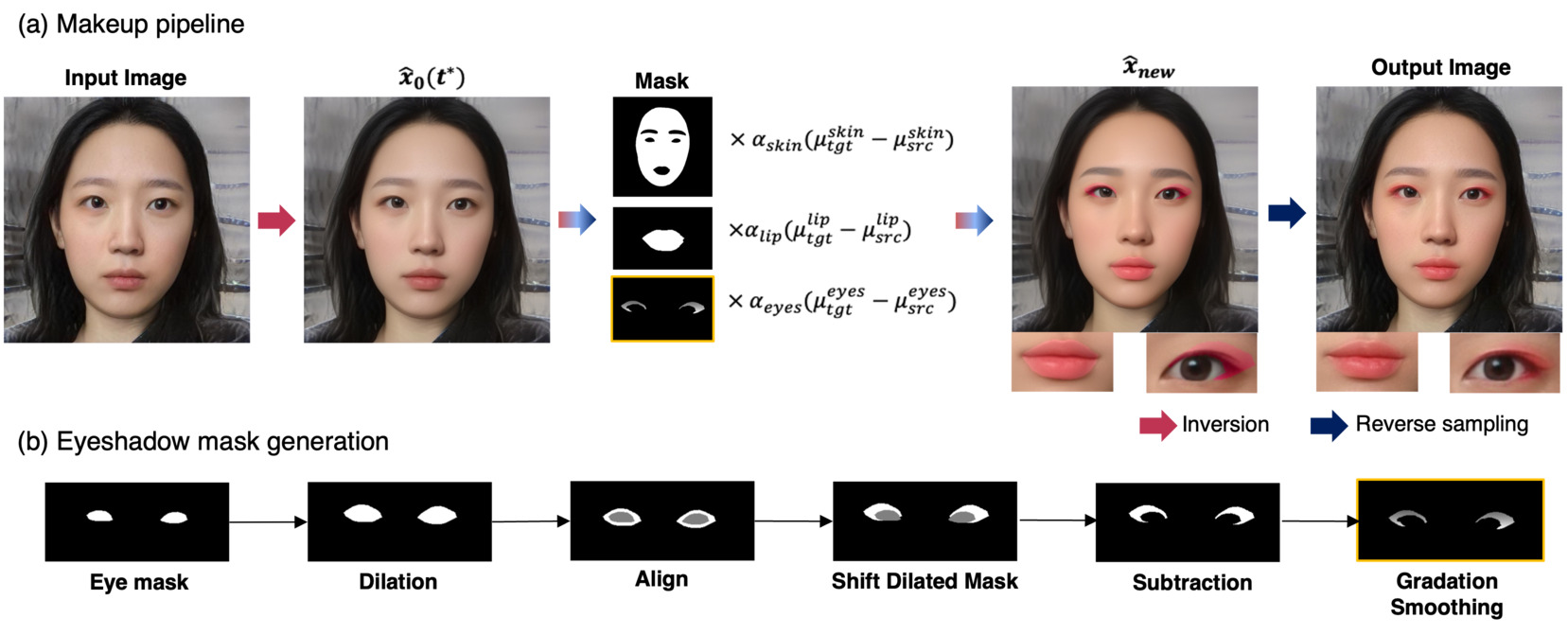}
    \caption{(\textbf{a}) Color-based makeup transformation. Mean RGB values within the masked area are adjusted with a scale $\alpha$ to match the target RGB values. Output image is generated by reverse sampling from $\hat{\xb}_{new}$. (\textbf{b}) Eyeshadow mask is reproduced from eye mask manipulation. 
    % (\textbf{c}) Makeup transfer and (\textbf{d}) Color-based makeup examples by DreamMakeup (SDXL). Each color chip represents lips, skin and eyeshadow.
    }
    \vspace{-0.4cm}
    \label{fig:RGB_method_figure}
\end{figure*}
\vspace{-0.2cm}

\begin{table}[t!]
\centering
\resizebox{\linewidth}{!}{
\begin{tabular}{@{\extracolsep{0pt}} c c c c c c}
\toprule
    {} & SDEdit & Ours & {} & SDEdit & Ours \\
\midrule
    LPIPS ($\downarrow$) & {0.069} & {\textbf{0.040}} & CLIP-I ($\uparrow$) & {0.868} & {\textbf{0.921}} \\
\bottomrule
\end{tabular}
}
\caption{Quantitative evaluation on preservation of facial structure.}
\vspace{-0.4cm}
\label{table: sdedit}
\end{table}

\subsection{Early-stopped Inversion}
%Latent Diffusion Model (LDM)\cite{rombach2022high} utilizes a latent domain approach rather than directly employing diffusion sampling within the image domain. First, the Encoder converts the image to a lower-dimensional latent variable. With DDIM forward sampling, noise is progressively added to the latent. The amount of forward sampling influences the reconstruction of the original image through the reverse process. 

While pixel-space editing achieves fine-grained control, the resulting edits must be seamlessly integrated without compromising the subject's identity. Standard harmonization techniques that rely on the \textit{stochastic} nature of the reverse diffusion process, often fail at this task by degrading the facial structure and identity. For instance, performing primary pixel-space makeup customization (\cref{sec: customization}) followed by noise addition and subsequent denoising refinement (SDEdit \cite{meng2022sdedit}) results in significant identity degradation (\cref{fig:sdedit}a-2, \cref{table: sdedit}). For successful makeup customization, the sampling trajectory must be constrained to preserve the original facial identity, as humans are highly sensitive to subtle facial changes. To faithfully maintain the identity of the original face during the sampling process, we instead leverage \textit{deterministic} DDIM inversion which is an iterative reverse simulation of the ODE flow in the limit of small steps. By setting $\eta=0$ in \cref{eq: ddim sampling}, the DDIM inversion \cite{mokady2023null, song2020denoising} for $\zb_t$ is defined as:
\begin{align}
    \zb_t = a_t \zb_{t-1} - b_t \epsilonb_{\theta} \big( \zb_{t-1}, t, c \big), 
\end{align}
where $a_t = \frac{\sqrt{\alphabar_t}}{\sqrt{\alphabar_{t-1}}}, b_t =  \sqrt{\alphabar_t} \Bigg( \sqrt{ \frac{1}{\alphabar_{t-1}} - 1} - \sqrt{ \frac{1}{\alphabar_t}-1 } \Bigg)$. 

This relies on the linearization approximation assuming $\epsilonb_{\theta} \big( \zb_{t-1}, t, c \big) \approx \epsilonb_{\theta} \big( \zb_t, t, c \big)$. However, the approximation error may accumulate following inversion steps, potentially causing identity loss. To prevent this, we terminate the inversion at $t^* \leq T$ to reduce computational burdens and ensure structural consistency, in contrast to the conventional process inverts $\zb_0$ to $\zb_T$. \cref{fig: overall} shows that the denoised $\hat{\zb}_0(t^*)$ from early-stopped $\zb_{t^*}$ may decode faithfully the original sample $\xb_0$. This allows us to directly guide $\hat{\xb}_0(t^*) = \Dc(\hat{\zb}_0(t^*))$ in a pixel space to enforce the target style. To further guarantee the faithful preservation, we propose interpolation-guided sampling which will be discussed in \cref{sec: EBCQ}.

%Note that we fix the textual condition $c$, e.g. \texttt{"a photo of a woman"}.
%With early-stopped forward samplings, the decoded image from latent space has a strong similarity with the original input face, and RGB color-based guidance and reference image guidance are applied in the decoded image. After makeup and makeup transfer with guidance, the image domain reverts to the latent space, and returns to its original domain with reverse process. The unnaturalness from makeup guidance is naturally addressed during the reverse process.  
% forward ODE by iterative simulation of the ODE flow

%\vspace{-0.3cm} 
\subsection{Pixel-space makeup customization}
\label{sec: customization}
Our next goal is to transform $\hat{\xb}_0(t^*)$ in a manner that accurately emulates the desired makeup appearance indicated by reference image or a target RGB color. To this end, we introduce a pixel-space transformation $\Tc(\cdot, \cdot): \mathbb{R}^{H \times W \times 3} \times X \rightarrow \mathbb{R}^{H \times W \times 3}$, offering multiple variants of $\Tc$ to enable fine-grained makeup customization. Here $X$ varies for different references, e.g. target RGB color, reference image, etc. 

%Further, diffusion models can approximate the denoised estimates $\mathbb{E}[\xb \given[] \xb_t]$ by using the denoiser pre-trained with Denoising Score Matching (DSM) \cite{vincent2011connection, song2019generative}. Starts from the inverted latent code $\xb_t$, this estimate is in the same dimensional space with the input face image and maintain overall facial structure. Therefore, these denoised estimates can be directly leveraged to the facial makeup.

\subsubsection{Makeup transformation with RGB color}
We first delineate an intuitive color transfer function that imposes the color characteristics of the reference makeup palette color on the source image. Let $\mu_{src}(\hat{\xb}_0(t^*))$, $\sigma_{src}(\hat{\xb}_0(t^*))$ represents the RGB mean and standard deviation of $\hat{\xb}_0(t^*)$ computed across spatial dimensions. Given a reference color $\mu_{tgt}$ and respective standard deviation $\sigma_{tgt}$, the color transfer function $\Tc_{RGB}$ is defined as 
\begin{align}
    \label{eq: RGB transformation}
    \hat{\xb}_{new}(t^*) &= \Tc_{RGB} \big(\mu_{src}(\hat{\xb}_0(t^*)), \mu_{tgt}; \alpha \big) \\
    &= \frac{\sigma_{src}(\hat{\xb}_0(t^*))}{\sigma_{tgt}} \Big(\hat{\xb}_0(t^*) - \alpha \big[\mu_{src}(\hat{\xb}_0(t^*)) - \mu_{tgt} \big] \Big), \nonumber
\end{align}
where $0 \leq \alpha \leq 1$ represents a transfer scale. 
For simplicity, we empirically set $\sigma_{src}(\hat{\xb}_0(t^*)) = \sigma_{tgt}$.

%\noindent\textbf{Color makeup composition.} 

The proposed color transfer supports attribute compositions. Specifically, we may transfer different $\mu_{tgt}$ for lips, eye shadow, skin foundation, etc. For this, we segment each interested facial attribute using a segmentation model \citep{yu2018bisenet} pre-trained in pixel domain. We observed that the inverted image $\hat{\xb}_0(t^*)$ is well segmented by the pre-trained model, owing to its high similarity with the original image $\xb_0$. 

%\textcolor{blue}{
%Makeup application requires fine-grained control over low-level visual attributes, such as color and edges. Direct manipulation in the pixel space can introduce color inconsistencies (\cref{fig:sdedit}a-1), yet these attributes are often nonlinearly encoded in the latent space, making direct latent-based customization challenging. To resolve this dilemma, we introduce an intermediate step that redirects the denoised estimates to the pixel domain. In this domain, low-level visual features can be more easily identified and manipulated. This technique of performing \textit{one-step}, fine-grained customization in the \textit{pixel domain} is a significant departure from existing \textit{multi-step} guidance methods that operate exclusively in the latent or attention spaces \citep{hertz2023delta, kwon2022diffusion, mokady2023null}.
%} 

To prevent potential color discontinuities and ensure seamless integration, the eyeshadow and lip masks undergo further refinement. The eyeshadow mask is first reconstructed from the initial eye mask using transformations such as dilation and shifting, after which its edges are smoothed to create a natural blend (\cref{fig:RGB_method_figure}). Specifically, this smoothing is achieved by applying a gradient mask whose weights increase progressively from the inner to the outer boundary. The lip masks are also refined using a similar smoothing process. This allows fine control, enabling users to adjust the gradient decay rate for a more natural appearance. Any artifacts from this color transfer process are further refined through reverse diffusion sampling (More details in \cref{sec: EBCQ}). 

%Specifically, the dilated mask shifts up to align the height with the original mask. Then, the overlapped area is subsequently subtracted to eliminate the lower eyelid, thereby isolating the upper section. Then, $\mu_{src}$ is obtained from the masked region, and the region is subsequently infused with the target RGB color on average. 
%ChatGPT: While the lips and facial masks are readily obtainable through segmentation, the eye shadow mask requires additional steps: it is produced by dilating and shifting the eye mask region. The original colors are extracted using these masks, which are then attenuated within the mask regions on the face. Subsequently, the RGB colors are applied to the masks, followed by edge smoothing to enhance naturalness. This process culminates with the reverse process, which serves to refine the resultant additions of RGB colors, mitigating any artificiality in the final image.

\subsubsection{Makeup transfer with reference image}
The transformation $\Tc$ can be varied depending on the downstream task. To demonstrate its universality, we consider conventional makeup transfer tasks \citep{li2018beautygan}. Specifically, we simulate the makeup style of reference image through warping and histogram matching transformations. 

First, histogram matching aligns the color distributions of the lip, eye shadow, and skin with reference. 
Then, the eyes of the source face are aligned with the reference through a series of warping transformations, including segmentation, dilation, affine, and diffeomorphic transformations, to ensure precise registration. This ensures that every pixel within the dilated mask area of the reference image corresponds to the appropriate region on the source image. 
% The eyeshadow mask is then transferred from the reference to the source with additional smoothing for natural edge transitions.
These steps enable the seamless adoption of tones and styles of the reference image, ensuring a natural makeup transfer.

\subsection{Makeup harmonization in reverse sampling}
\label{sec: EBCQ}
After pixel-space guidance, the output $\hat{\xb}_{new}$ coarsely follows the desired makeup style in the local facial attributes, e.g. lips, eye shadow, etc, but could suffer from unnatural appearances ($\hat{\xb}_{new}$ as in \cref{fig:RGB_method_figure}), requiring further refinement. Thus, our next goal is to (\textbf{a}) stylize $\hat{\xb}_{new}$ with \textit{multiple} text descriptions to harmonize such local variations with a consistent aesthetic style, e.g. ``Korean K-Pop style", ``fair skin", etc, while (\textbf{b}) maximally preserving the facial identity. This refinement begins with $\zb_t = \sqrt{\alphabar_t} \Ec(\hat{\xb}_{new}(t^*)) + \sqrt{1-\alphabar_t}\epsilonb(\zb_{t^*}, t^*, c)$, followed by subsequent denoising.

\noindent\textbf{(a): Cross attention composition.} Precise control over diverse makeup demands and facial attributes is challenging due to the need to balance multiple prompt strengths. To address this, we refine the image by integrating semantic makeup features directly into the cross-attention layer.
Specifically, let $\Q_{t,l} \in \Rb^{P_l^2 \times  d_l }$ represent the spatial query in $l$-th cross attention layer of U-Net. Given context vectors $\Cb \in \mathbb{R}^{N \times d_c}$, let $\K, \V \in \Rb^{dc \times d_l}$ denote key and value matrices, respectively, where $N$ refers to number of tokens, $\K= \C \W_{K, l}$ and $\V= \C \W_{V, l}$ with linear maps $\W_{K, l}, \W_{V, l} \in \Rb^{d_c \times d_l}$. Then, let $\K_s$ represents a $s$-th makeup concept key $\K_s$, where $\K_{main}$ comes from the prompt used in inversion, i.e. ``a photo of a woman". Define $\V_s, \V_{main}$ similarly. Then, we update the spatial query as:
\vspace{-0.1cm}
\begin{align}
\label{eq: composition E}
\Q^{new} &= \softmax \big(\frac{\Q \K_{main}^T}{\sqrt{d}} \big) \V_{main} + \\
&\frac{1}{M} \sum_{s=1}^{M} \alpha_s \softmax \big(\frac{\Q \K_{s}^T}{\sqrt{d}} \big) \V_s.
\end{align}
The degree and direction of $s$-th makeup concept can be controlled \textit{individually} with $\alpha_s$, where $\alpha_s<0$ for negative makeup prompts, e.g. \texttt{ugly, blurry, low-res}. This linear combination incorporates detailed makeup descriptions independently, allowing fine-grained control of prompt strengths. %We empirically observed that the proposed sampling method works well even without external regional masks, which are commonly used in diffusion-based customization frameworks \citep{gu2024mix, kwon2024concept}. This may be attributed to the global nature of makeup prompts focusing on aesthetic style, and the separation of facial attributes within semantic features. Overall pipeline is summarized in \cref{fig: overall} and pseudo-code in appendix.

\noindent\textbf{(b): Interpolation-guided sampling.} While cross-attention composition effectively steers ODE sampling toward desired makeup styles, facial identity may degrade due to accumulated errors in DDIM inversion. As shown in \cref{fig:sdedit}b, DDIM inversion may fail to preserve fine-grained skin texture (e.g., pores, micro-blemishes), leading to an unnatural, \textit{plastic} skin--an issue widely recognized in the SD community. To mitigate this, we \textit{regularize} the sampling trajectory to remain closer to the original facial latents:
\vspace{-0.1cm}
\begin{align}
    \tilde{\zb}_0(t) &= \arg \min_z \| \zb - \hat{\zb}_0(t) \|^2 + \frac{\lambda}{1-\lambda} \| \zb - \Ec (\Tc(\xb_0)) \|^2  \nonumber \\ 
                    &= (1-\lambda) \hat{\zb}_0(t) + \lambda \Ec (\Tc(\xb_0)), \\
    {\zb}_{t-1} &= \sqrt{\alphabar_t} \tilde{\zb}_0(t) + \sqrt{1-\alphabar_t} \epsilonb_\theta(\zb_t, t, c), \nonumber
\end{align}
where $\zb_0' = \Ec (\Tc(\xb_0))$ represents the makeup-transformed input latent. Serving as a pivotal reference, $\zb_0'$ helps retain photorealistic skin details, ensuring both realism and identity preservation (\cref{fig:sdedit}b-1) while maintaining the intended makeup from $\Tc(\cdot)$. Conversely, performing this interpolation directly in the pixel space via $(1-\lambda) \Dc (\hat{\zb}_0(t)) + \lambda \Tc(\xb_0)$ introduces undesirable visual artifacts, as shown in \cref{fig:sdedit}b-2. We observe that the latent space is a more suitable domain for this operation, facilitating a seamless interpolation between the denoised estimate and the transformed source. This approach strikes a superior balance between preserving the source's identity and applying the custom makeup requirements. Notably, applying this regularization in just 1–2 early sampling steps significantly enhances identity preservation, with a moderate level of $\lambda=0.15$.

\begin{figure}[t!]
    \centering
    \includegraphics[width=0.9\linewidth]{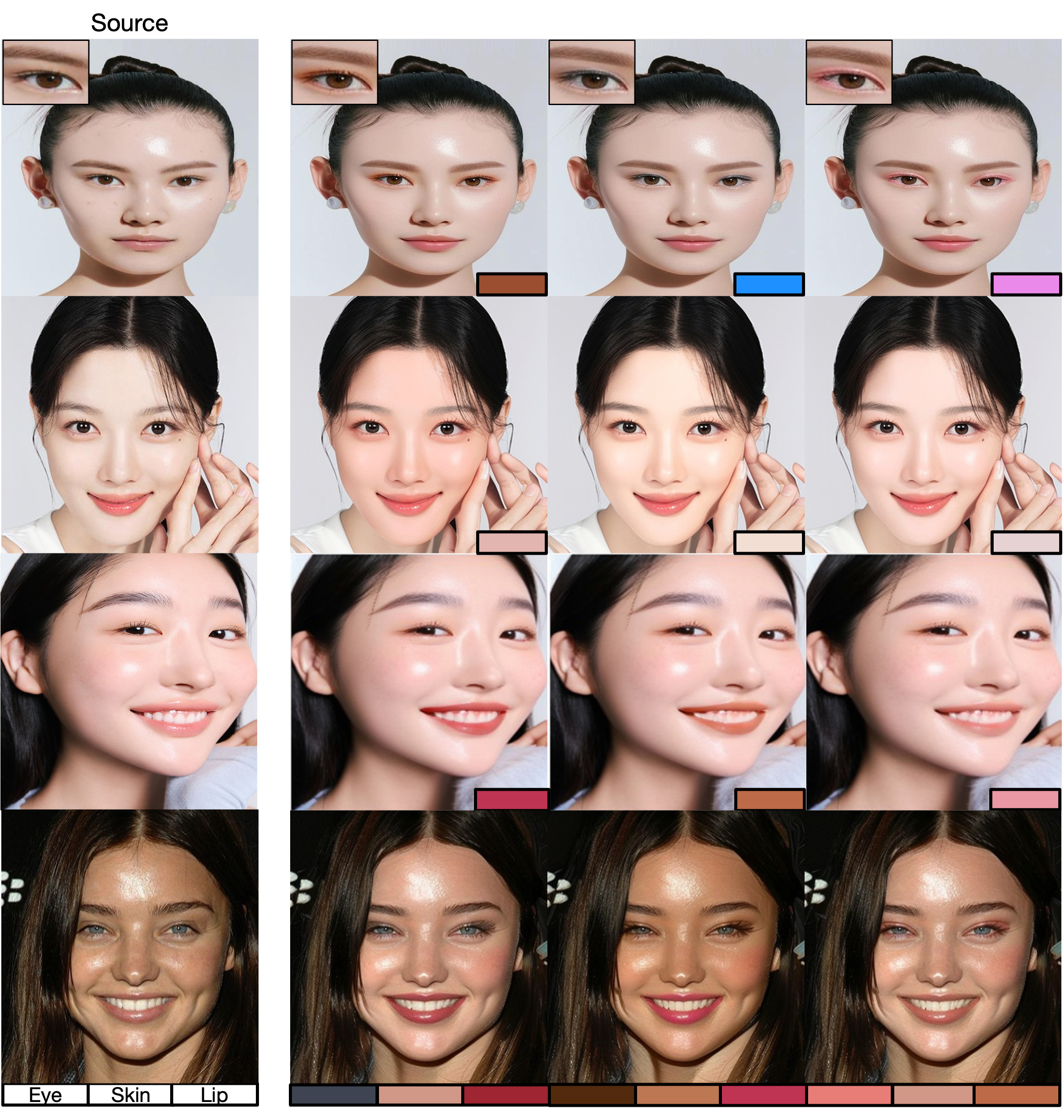}
    \caption{The virtual skin, lip, eye shadow makeup, and their combination by DreamMakeup (SD 1.5).}
    \vspace{-0.4cm}
    \label{fig:RGB_composition_opendata}
\end{figure}

\begin{figure*}[t!]
    \centering
    \includegraphics[width=\textwidth]{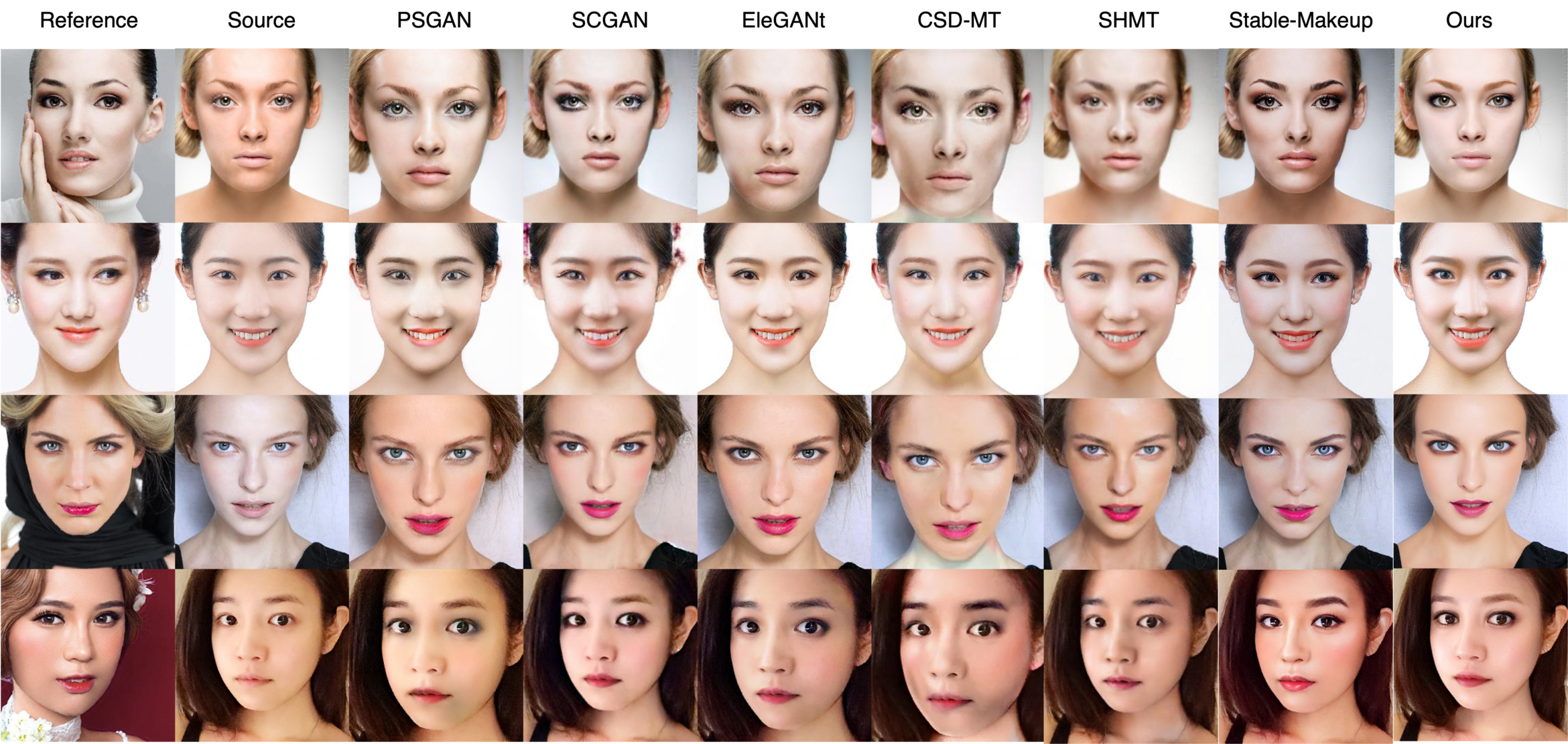}
    \caption{Qualitative comparison of reference image based makeup using references with diverse poses and makeup styles. GAN-based models exhibit artifacts and cropping issues (row 1,3,4). SHMT incorrectly transfers reference color and Stable-Makeup struggles with identity preservation, likely due to the stochastic nature of diffusion models.
    }
    \vspace{-0.3cm}
    \label{fig:Reference_comparison}
\end{figure*}

\section{Experimental Results}
%The proposed method utilizes finetuning model and LoRA\cite{hu2021lora} for makeup. data, control experiment

\subsection{Experiment Settings}
During inference, we used the Makeup Transfer (MT) dataset \cite{li2018beautygan} for both source and reference images. Additionally, we employed artificially generated images for Asian women as a non-makeup source image. We used Stable Diffusion (SD) v1.5 and SDXL \citep{podell2023sdxl} as our base model, and further leveraged additional public LoRA weights and pre-trained models released in CivitAI, an open-source generative AI community. For facial segmentation, we utilized BiSeNet \cite{yu2018bisenet}. For cross attention composition, we set the scaling factor for each makeup concept $\alpha_s$ ranging from $0.1$ to $0.7$. For comparison, we tested state-of-the-art GAN-based makeup transfer methods, PSGAN \cite{jiang2019psgan}, SCGAN \cite{deng2021spatially}, EleGANt \cite{yang2022elegant}, and CSD-MT \cite{sun2024content}, alongside SHMT \cite{sun2024shmt} and Stable Makeup \cite{zhang2024stable}, which are recent diffusion-based makeup transfer frameworks. More experimental details are in appendix.

% experiment parameters (ex. max_timestep, num_steps, alpha, editing prompt, lora scale, )
%In all experiments, we use DDIM scheduler and set the early-stop inversion step as $t^*=300$ and the number of reverse steps as $30$. LoRA scale $s$ is set to $0.2$: $W_{\text{new}} = W_0 + s * \triangle W_{\text{LoRA}}$, where we apply LoRA mainly to pre-trained weights $W_0$ of every attention layer of U-Net and textual encoder. We conduct the experiments on a NVIDIA RTX 3090 GPU. 

%\begin{figure}[t!]
%    \centering
%    \includegraphics[width=\linewidth]{Figure/text_guided_singlecolumn.jpg}
%    \caption{\textbf{Text-guided makeup transformation}. Top row: text-guided makeup transformation. Bottom row: text %+ RGB color guided transformations.}
%    \label{fig:text_guided}
%\end{figure}

\begin{table}[htbp]
\begin{subtable}[t]{0.48\textwidth}
\begin{adjustbox}{width=\linewidth,center}
\centering
\begin{tabular}{ccccccc}
\toprule
\multirow{2}{*}{\textbf{Method}} & \multirow{2}{*}{LPIPS $\downarrow$} & \multirow{2}{*}{CLIP-I $\uparrow$} & \multicolumn{2}{c}{Makeup Artists} & \multicolumn{2}{c}{Non Artists} \\
\cmidrule(rl){4-5} \cmidrule(rl){6-7}
{} & {} & {} & {Detail $\uparrow$} & {Quality $\uparrow$} & {Detail $\uparrow$} & {Quality $\uparrow$} \\
\midrule
PSGAN & 0.1879 & 0.7421 & 1.49 & 1.74 & 2.64 & 2.82  \\
SCGAN & \underline{0.0819} & 0.7253 & 2.00 & 2.11 & 3.03 & 2.95 \\
EleGANt & 0.1877 & \underline{0.7662} & \underline{3.04} & \underline{3.12} & \underline{3.67} & \underline{3.72} \\
Ours & \textbf{0.0667} & \textbf{0.7694} & \textbf{3.42} & \textbf{3.42} & \textbf{3.95} & \textbf{4.00} \\
\bottomrule
\end{tabular}
\end{adjustbox}
\captionsetup{width=\linewidth}
\caption{Quantitative results on makeup transfer tasks.}  
\label{tab:result_userstudy}
\end{subtable}
\hspace{\fill}
\begin{subtable}[t]{0.48\textwidth}
\begin{adjustbox}{width=\linewidth,center}
\centering
\begin{tabular}{cccccc}
\toprule
\multirow{2}{*}{\textbf{Method}} & \multirow{2}{*}{\makecell{Beauty \\ score}} & \multicolumn{2}{c}{Makeup Artists} & \multicolumn{2}{c}{Non Artists} \\
\cmidrule(rl){3-4} \cmidrule(rl){5-6}
{} & {} & {Detail $\uparrow$} & {Quality $\uparrow$} & {Detail $\uparrow$} & {Quality $\uparrow$} \\
\midrule
Service A & \underline{2.90} & \underline{4.08} & 1.97 & \underline{3.69} & 2.77 \\
Service B & 3.27 & 3.75 & \underline{2.61} & 3.14 & \underline{2.83}  \\
Ours & \textbf{3.38} & \textbf{4.22} & \textbf{4.19} & \textbf{3.93} & \textbf{4.27} \\
\bottomrule
\end{tabular}
\end{adjustbox}
\captionsetup{width=\linewidth}
\caption{Comparisons with global AI makeup services.}  
\label{tab:RGB}
\end{subtable}
\caption{Quantitative comparisons on makeup transfer task and color-based makeup transformation.}
\vspace{-0.5cm}
\end{table}

\subsection{Results}

\subsubsection{Qualitative Results}
\cref{fig:RGB_composition_opendata} illustrates the RGB color-based makeup transformations applied to both synthetic and natural images. First three rows reflect the application of the eye shadow, skin, and lip colors indicated in the bottom right corner. Bottom row presents the combined results for each column's corresponding colors. DreamMakeup effectively applies the specified RGB colors to the respective facial regions.

Qualitative comparison with reference-based makeup baselines is presented in \cref{fig:Reference_comparison}. Our approach demonstrates superior performance over competing models. Most GAN-based methods - PSGAN, SCGAN, and CSD-MT produce noticeable artifacts, including identity loss and color bleeding. While EleGANt performs competitively, it fails on challenging cases, exhibiting dark artifacts in occluded regions (row 1) or generating weaker eye makeup. Among diffusion-based methods, SHMT struggles with inaccurate color transfer, and Stable Makeup exhibits poor identity preservation. Our method consistently produces clean, artifact-free results that faithfully replicate the reference makeup style.

\begin{figure}[t!]
    \centering
    \vspace{-0.4cm}
    \includegraphics[width=\linewidth]{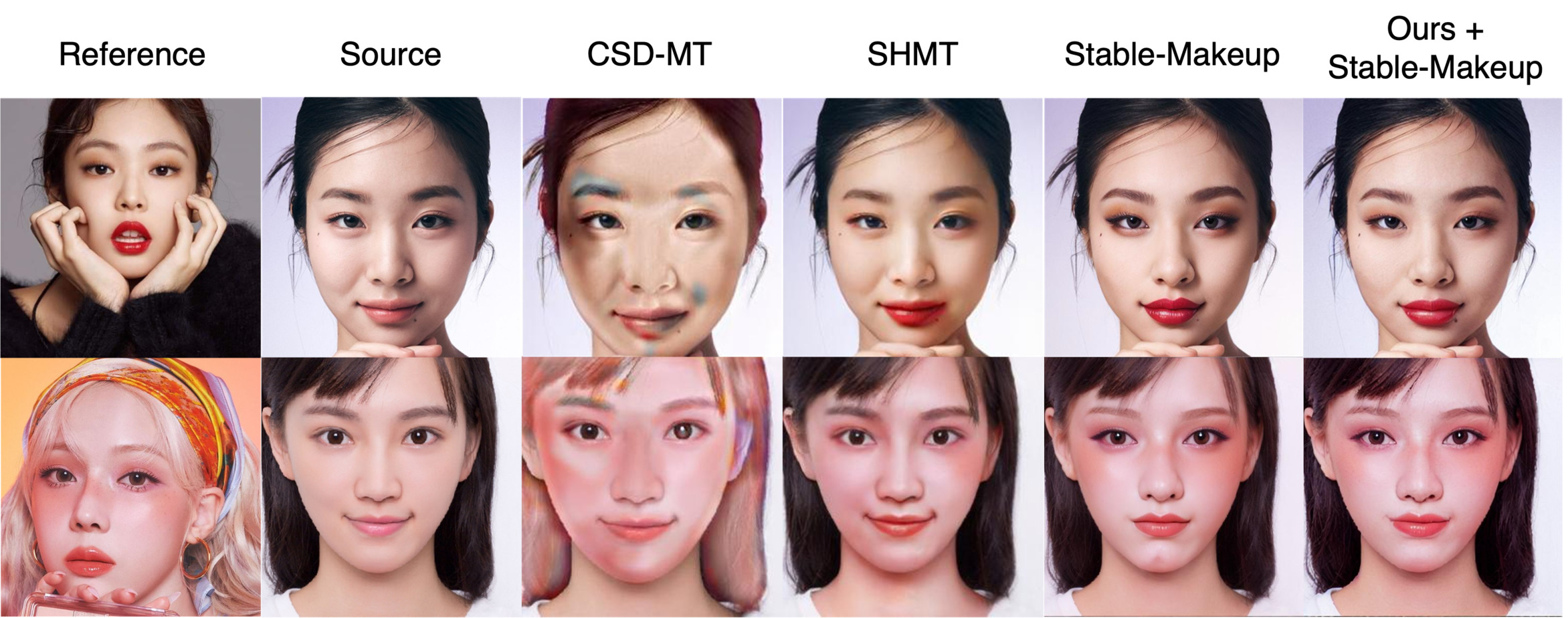}
    \caption{DreamMakeup integrated with Stable Makeup \cite{zhang2024stable}.}
    \label{fig: stable makeup}
    \vspace{-0.4cm}
\end{figure}

\begin{figure}[t!]
    \centering
    \includegraphics[width=\linewidth]{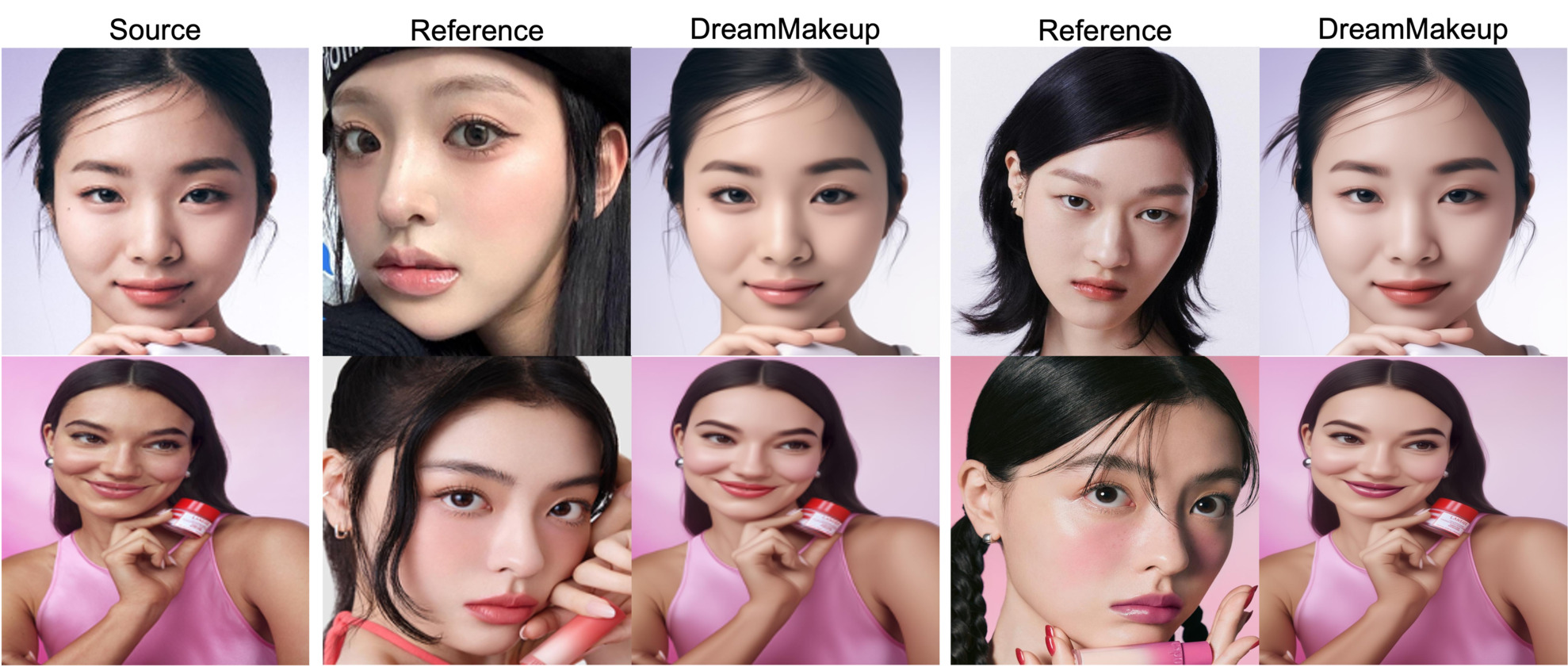}
    \caption{DreamMakeup integrated with Flow-based models.
    }
    \vspace{-0.4cm}
    \label{fig:dit}
\end{figure}

\begin{figure}[t!]
    \centering
    \includegraphics[width=0.9\linewidth]{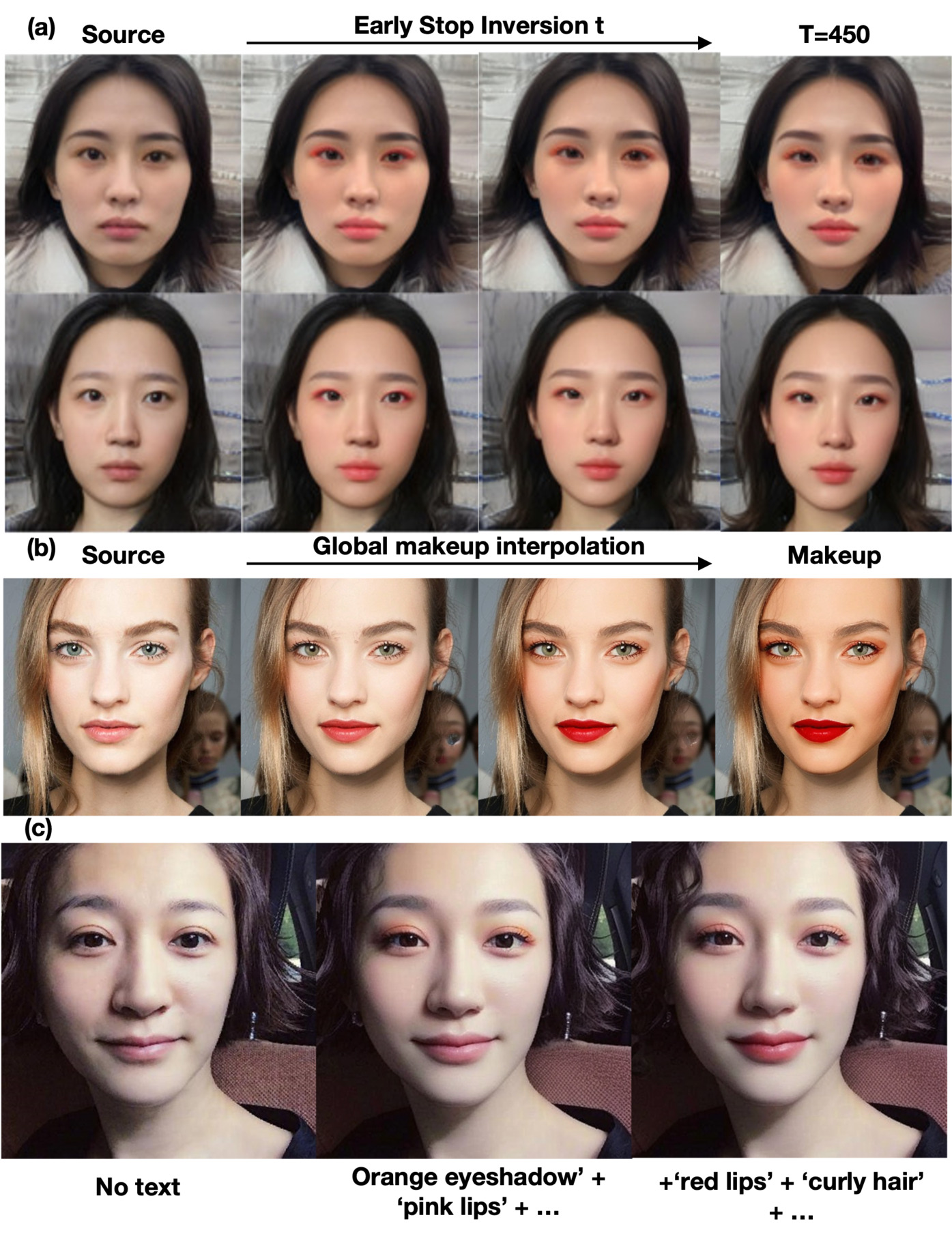}
    \caption{Ablation study for early-stop timestep $t$, interpolation scale $\alpha$ and text guidance.}
    \label{fig:ablation}
    \vspace{-0.4cm}
\end{figure}

As a training-free framework, DreamMakeup can be seamlessly integrated with other generative models. We demonstrate this modularity by incorporating our core components such as inversion, pixel-space customization, and interpolation guidance, into two backbones: the diffusion-based makeup transfer framework Stable Makeup \cite{zhang2024stable} and the recent flow-based Stable Diffusion 3.0 \cite{esser2024scaling}. In \cref{fig: stable makeup}, integrating DreamMakeup improves structural consistency and reduces artifacts compared to the original Stable Makeup model. Furthermore, \cref{fig:dit} illustrates that our framework is compatible with Diffusion Transformer architectures, highlighting its potential for synergistic applications across different model families.
 
\subsubsection{Quantitative Comparison}
We evaluate DreamMakeup against baselines on both color-based and reference image-based makeup tasks. For reference based makeup, we assess identity preservation using LPIPS and CLIP image similarity. Additionally, 10 expert makeup artists and 24 non-experts ranked 10 random outputs based on style accuracy and overall quality (scale: 1–5), where DreamMakeup consistently excelled (\cref{tab:result_userstudy}).

For color-based makeup, we further compare against two global AI makeup services (50M+ downloads) while ensuring anonymity. Despite their widespread use, these services exhibit limitations in customization, such as restricted color presets. Using 100 images per service, we evaluate beauty scores \citep{xu2019hierarchical} and conduct a broader user study (300 images). DreamMakeup consistently outperforms others (\cref{tab:RGB}).
\vspace{-0.1cm}

\begin{figure}[htb]
    \centering
    \includegraphics[width=\linewidth]{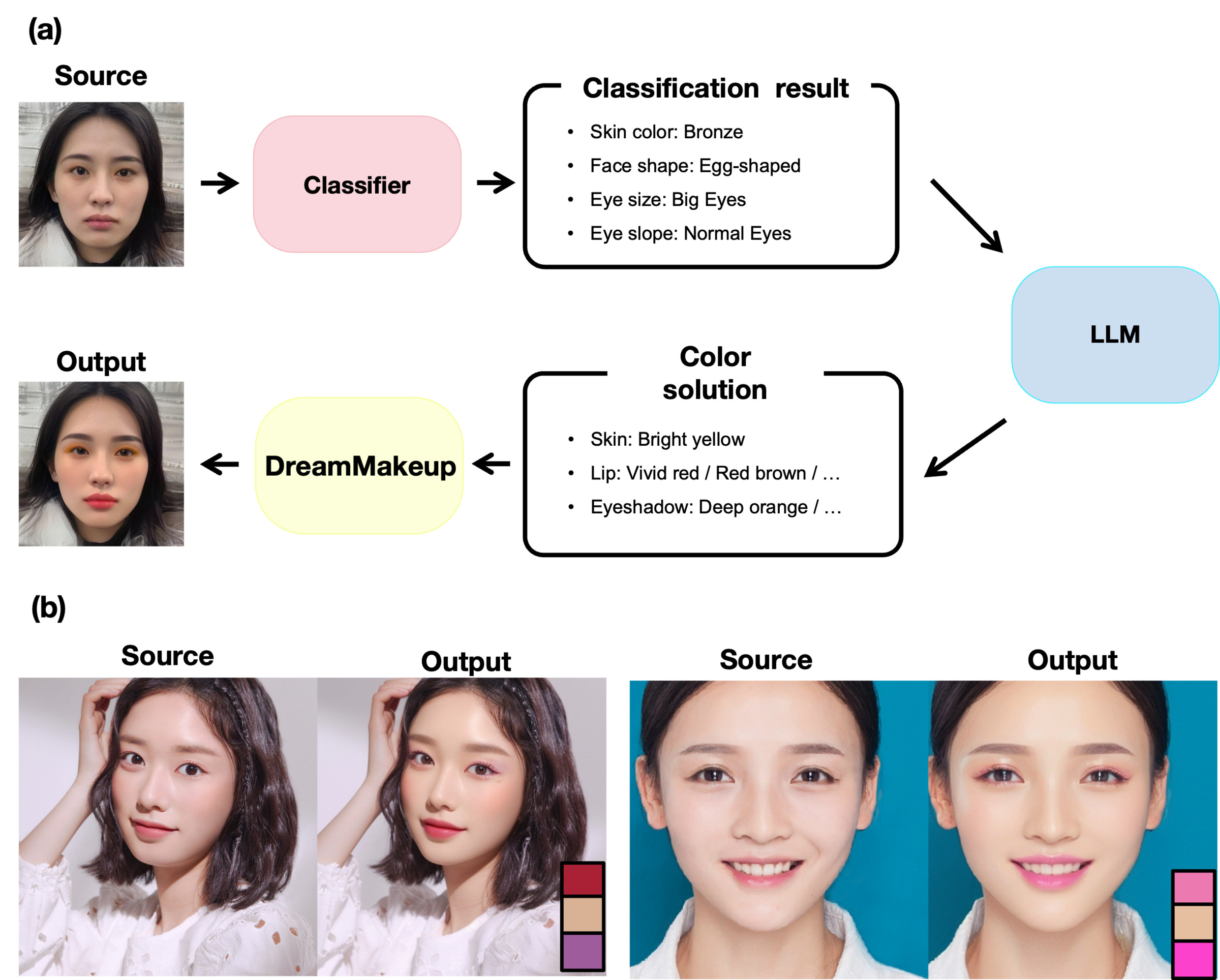}
    \caption{(a) DreamMakeup with integration with a classifier and an LLM. (b) From the source image, we can apply makeup based on the solution provided by the classifier and LLM. The color chips represent the color of lips, skin, and eye shadow from top to bottom.}
    \vspace{-0.5cm}
    \label{fig:pipeline_llm}
\end{figure}

%\subsubsection{Text-guided makeup transformation}
%To demonstrate the efficacy of text guidance, we apply makeup on the source image with (1) only using text guidance with cross attention composition, and (2) using text guidance along with RGB color transfer. \cref{fig:text_guided} illustrates that text guidance effectively facilitates makeup transformation, and concisely stylizes the RGB-based makeup transformations. Specifically, the color transfer in the second column establishes the overall color distribution, and linguistic conditioning further improves the natural appearance.

\subsubsection{Ablation Study}
We conducted an ablation study on the key hyperparameters, with results shown in \cref{fig:ablation}. Varying the early-stop timestep $t^*$ for DDIM inversion allows for a balance between faithful identity preservation and stylistic variation. The transfer scale $\alpha$, defined in \eqref{eq: RGB transformation}, governs the intensity of the applied makeup. As $\alpha$ approaches 0, the output progressively resembles the original source image, effectively reducing the makeup's opacity (\cref{fig:ablation}(b)). Notably, this parameter can be adjusted independently for distinct facial regions (e.g., lips, eyes), enabling fine-grained control over the final look. The effect of textual guidance, implemented via cross-attention composition, is shown in \cref{fig:ablation}(c). The inclusion of descriptive prompts harmonizes the local makeup applications with a cohesive global aesthetic, demonstrating the significance of text conditioning in achieving a polished result.

%The classifier analyzes the input image to assess skin tone, face shape, eye size, eye shape, and more. Then, based on these diagnostic results, the language model suggests makeup colors that complement the input image. Finally, our model applies makeup to the input image based on these colors. 

\subsection{Integration with LLM}
DreamMakeup can  demonstrate improved performance by integrating with Large Language Models (LLMs) for personalized makeup recommendations. By harnessing the exceptional inference capabilities of LLMs, DreamMakeup  selects makeup colors that are harmonious with the characteristics of the source image. We provide a pipeline in \cref{fig:pipeline_llm}. This pipeline involves an initial extraction of facial attributes such as skin tone and facial structure from the source image via a classifier. This information is then conveyed to the LLM, which determines the most appropriate makeup colors for various facial regions, including the skin, eyes, and lips. DreamMakeup subsequently utilizes these recommendations to generate the final makeup-enhanced image.

These components (classifier, LLM, and DreamMakeup) operate sequentially during inference but are trained independently. The classifier was trained using ResNet50 on a dataset of 1,000 artificially generated images of Asian women, annotated with facial information. The LLM is trained on the dolly-v2-3b model with a specialized QnA dataset from beauty professionals. As illustrated in \cref{fig:pipeline_llm}(b), the LLM adeptly matches skin, eye shadow, and lip colors to the source image, facilitating the application of these colors by DreamMakeup. The process ensures that the selected colors are well-suited to the source image, leading to an effectively applied makeup look. This demonstrates the potential of integrating classifiers and LLMs in DreamMakeup to provide customized makeup solutions based on in-depth analysis of facial features, thereby enhancing the personalization and effectiveness of makeup applications.

\section{Conclusion}
We introduced DreamMakeup, a novel training-free makeup framework utilizing powerful priors. Our method ensure precise makeup customization with rich conditions. We demonstrated DreamMakeup's global effectiveness and efficiency in various tasks and verify compatibility with other frameworks.

{
    \small
    \bibliographystyle{ieeenat_fullname}
    \bibliography{main}
}

\newpage

\appendix

\section{Pseudo Code}
We provide the pseudo-code of DreamMakeup for RGB and textual guidance in \cref{alg:DreamMakeup}. Makeup transfer based on a reference image can be easily implemented using the same method, substituting the transformation $\Tc_{RGB}$ with $\Tc_{ref}$ and employing warping and histogram matching algorithms instead of RGB matching.

\begin{algorithm}[h]
\caption{DreamMakeup with RGB and text guidance}\label{alg:DreamMakeup}
\textbf{Input:} Source image $\xb_0$, early-stop timestep $t^* \le T$, RGB scaling coefficient $0 \le \alpha \le 1$, target RGB color $\mu_{tgt}$, reference makeup image $\xb_{ref}$, textual prompts for (inversion, editing) $C_{inv}$, $\{C_{edit, s}\}_{s=1}^N$, degree of composition $\{\alpha_s\}_{s=1}^N.$ \\
\textbf{Output:} Image with makeup transformation $\tilde{x}_0.$
\begin{algorithmic}[1]
    \State $\zb_0 = \Ec(\xb_0)$ \\
    
    \State \textit{1. Early-stopped DDIM inversion}
    \For {$t=1$ \textbf{to} $t^*$}
        \State $\zb_t = \text{DDIM-Inv}(\zb_{t-1}, t, c)$  (Sec. 4.1)
        %\State $\sqrt{\alphabar_t} \Bigg( \sqrt{ \frac{1}{\alphabar_{t-1}} - 1} - \sqrt{ \frac{1}{\alphabar_t}-1 } \Bigg) \epsilonb_{\theta} \big( \zb_{t-1}, t, c) \big)$.
    \EndFor
    
    \State $\hat{\zb}_0(t^*) = \frac{1}{\sqrt{\alphabar_{t^*}}} (\zb_{t^*} - \sqrt{1-\alphabar_{t^*}} \epsilonb_{\theta} (\zb_{t^*}, t^*))$.
    \State $\hat{\xb}_0(t^*) = \Dc(\hat{\zb}_0(t^*))$ \\

    \State \textit{2. Pixel-domain Diffusion Guidance}
    \State $\hat{\xb}_{new} = \Tc_{RGB} \big(\mu_{src}(\hat{\xb}_0(t^*)), \mu_{tgt}; \alpha \big)$ 
    \State $\tilde{\zb}_{t^*} = \sqrt{\alphabar_t} \Ec(\hat{\xb}_{new}) + \sqrt{1-\alphabar_t}\epsilonb(\zb_{t^*}, t^*, c) $ \\
    
    \State \textit{3. Reverse sampling with cross attention composition}
    \For {$t=t^*$ \textbf{to} $1$}
        \State $\tilde{z}_{t-1} \gets \text{ReverseDDIM} \Big($ \par
        \hskip\algorithmicindent $\tilde{z}_{t}; t, \text{Composition} \big( \{\alpha_s\}_{s=1}^N, \{C_{edit, s}\}_{s=1}^N$ \big)\Big)
    \EndFor

\State $\tilde{x}_0 = \Dc(\tilde{z}_{0})$
\end{algorithmic}
\vspace{-0.1cm}
\end{algorithm}

\section{Additional analysis}
\begin{figure*}[t!]
    \centering
    \includegraphics[width=0.95\textwidth]{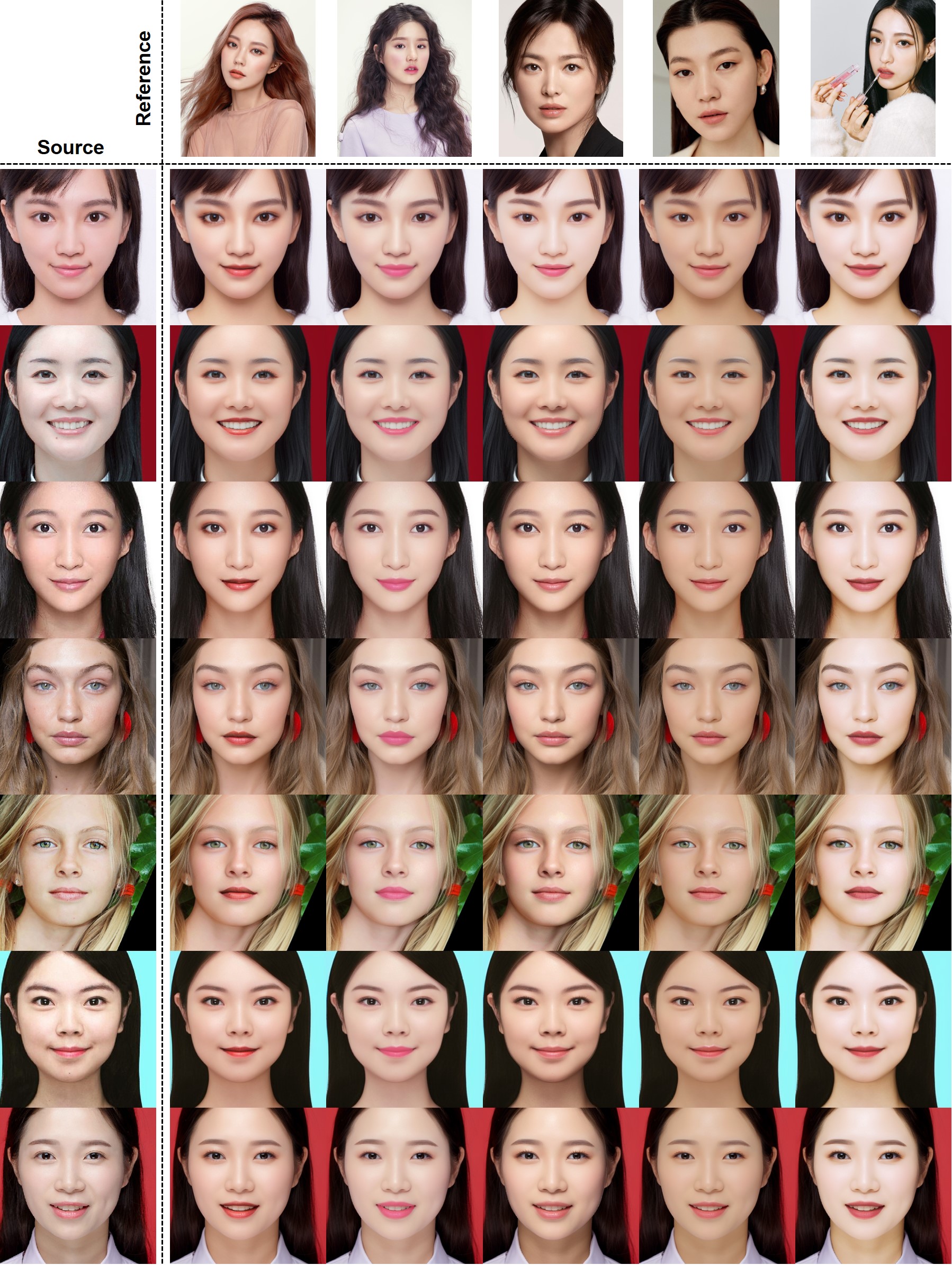}
    \caption{Addition results on makeup transfer.}
    \label{fig:supp makeup transfer}
\end{figure*}

\vspace{-0.1cm}
We provide in-depth analysis on the core components of DreamMakeup. Additional qualitative results are also provided in \cref{fig:supp makeup transfer}.
\subsection{Ablation studies}
\begin{figure}[t!]
    \centering
    \includegraphics[width=\columnwidth]{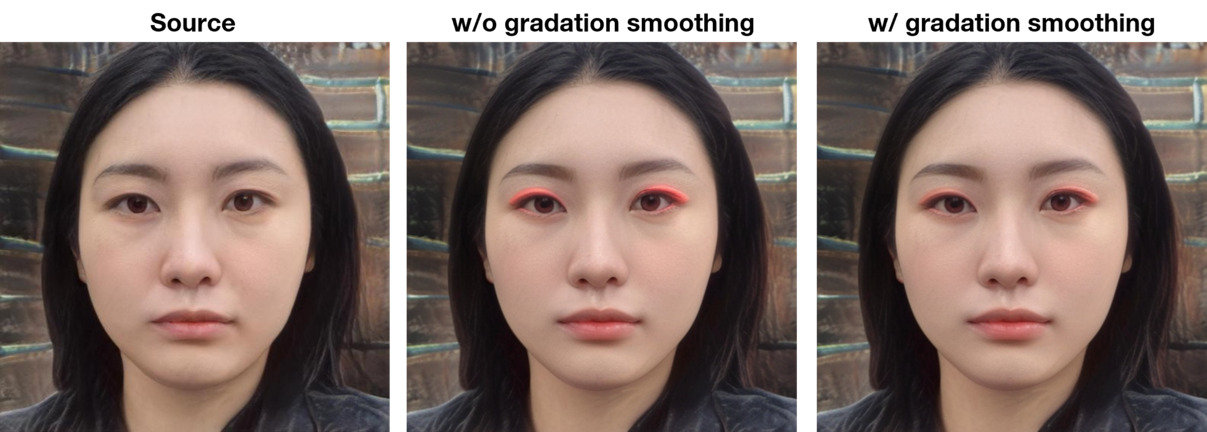}
    \caption{The effect of the gradation smoothing of eyeshadow mask. Please zoom in for detailed inspection.}
    \vspace{-0.2cm}
    \label{fig:effect_of_gradation}
\end{figure}

\textbf{Effects of gradation smoothing.} In the generating process of eye mask, gradation smoothing is essential. Without gradation smoothing, the edges of eye masks are accentuated, resulting in an unnatural outcome. \cref{fig:effect_of_gradation} demonstrates that graduation smoothing makes the edge of the eye shadow natural and realistic.

%\subsubsection{Effects of early-stopped DDIM inversion} 
%To assess the significance of each stage in the makeup pipeline, we visualize the intermediate results in \cref{fig:inversion output}. Specifically, we present $\hat{\xb}_0(t^*)$ (after inversion) and $\hat{\xb}_{new}$ (after color transformation). DDIM inversion retains the subject's identity, while the color transformation defines the overall makeup layout. Reverse sampling further harmonizes the virtual try-on makeup seamlessly.

%\cref{fig:effect_of_early_stop} demonstrates that early-stopping inversion provides a valuable knob for adjusting RGB makeup transformation fidelity and naturalness. Increasing $t^*$ improves target makeup representation at the affordable cost of higher computational demands, while decreasing $t^*$ preserves subject identity and ensures accurate color representation. This approach, including adjustments with $t^*$ and other parameters like $\alpha$, offers a remarkable customization capacity unavailable in conventional frameworks.

\begin{figure}[t!]
    \centering
    \includegraphics[width=\columnwidth]{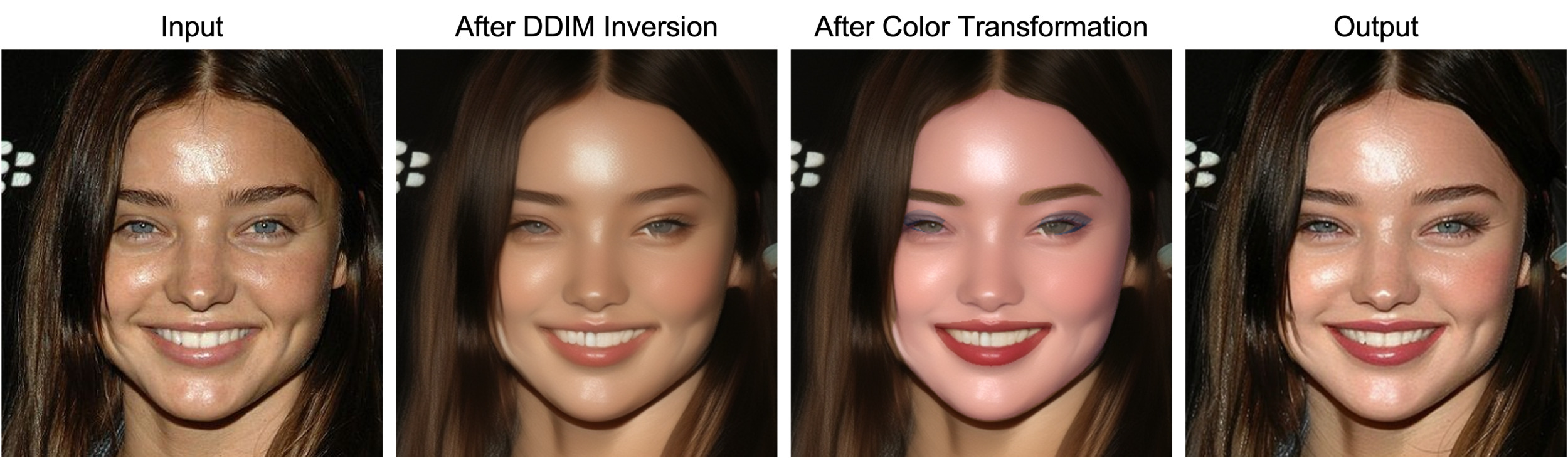}
    \caption{Ablation study on DDIM inversion, coloring, and reverse sampling. Text prompts guide the harmonization of unnatural color transitions into a cohesive aesthetic style during sampling.}
    \label{fig:inversion output}
    \vspace{-0.3cm}
\end{figure}

%\begin{figure*}[t!]
%    \centering
%    \includegraphics[width=0.9\textwidth]{Figure/early_stop_inversion.jpg}
%    \caption{The effect of the early-stopped DDIM inversion step $t^*$.}
%    \vspace{-0.3cm}
%    \label{fig:effect_of_early_stop}
%\end{figure*}

\subsection{LoRA variation}
\begin{figure}[t!]
    \centering
    \includegraphics[width=\columnwidth]{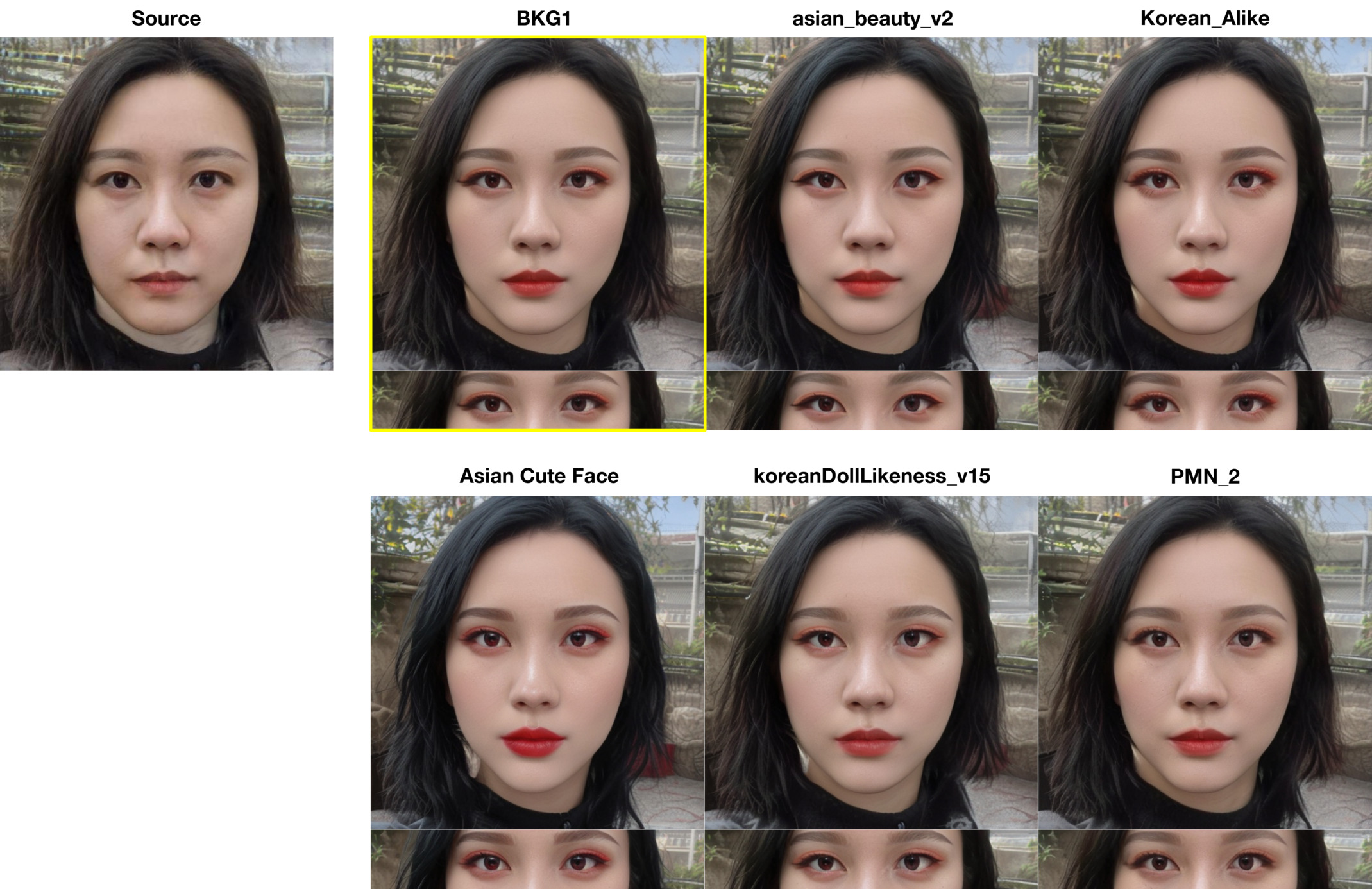}
    \caption{Results of using various LoRA weights.}
    \vspace{-0.2cm}
    \label{fig:lora_variation}
\end{figure}

We mainly utilized Dreamshaper\footnote{https://civitai.com/models/4384/dreamshaper}, ArienMixXL \footnote{https://civitai.com/models/118913/sdxl-10-arienmixxl-asian-portrait}, and BKG1\footnote{https://civitai.com/models/203947/beautiful-korean-girl-bkgv1} LoRA weights. \cref{fig:lora_variation} shows the experimental results of using other LoRA weights. For comparison, asian beauty v2\footnote{https://civitai.com/models/76883/2731-pretty-asian-face-asian-beauty-faces}
, Korean Alike\footnote{https://civitai.com/models/193777/korean-alike-by-noerman}, Asian Cute Face\footnote{https://civitai.com/models/26914?modelVersionId=32215}, koreanDollLikeness v15\footnote{https://civitai.com/models/26124/koreandolllikeness-v20}, PMN 2\footnote{https://civitai.com/models/106028/korean-beauty} are used. The results are made with only text guidance, where the prompts are \texttt{"deep red lip"} and \texttt{"heavy eye makeup"}. The results demonstrate how diverse makeup styles can be achieved by varying LoRA weights. In this paper, we mainly leverage BKG1 LoRA which shows better identity preservation and semantic alignment.

\begin{figure}[t!]
    \centering
    \includegraphics[width=\columnwidth]{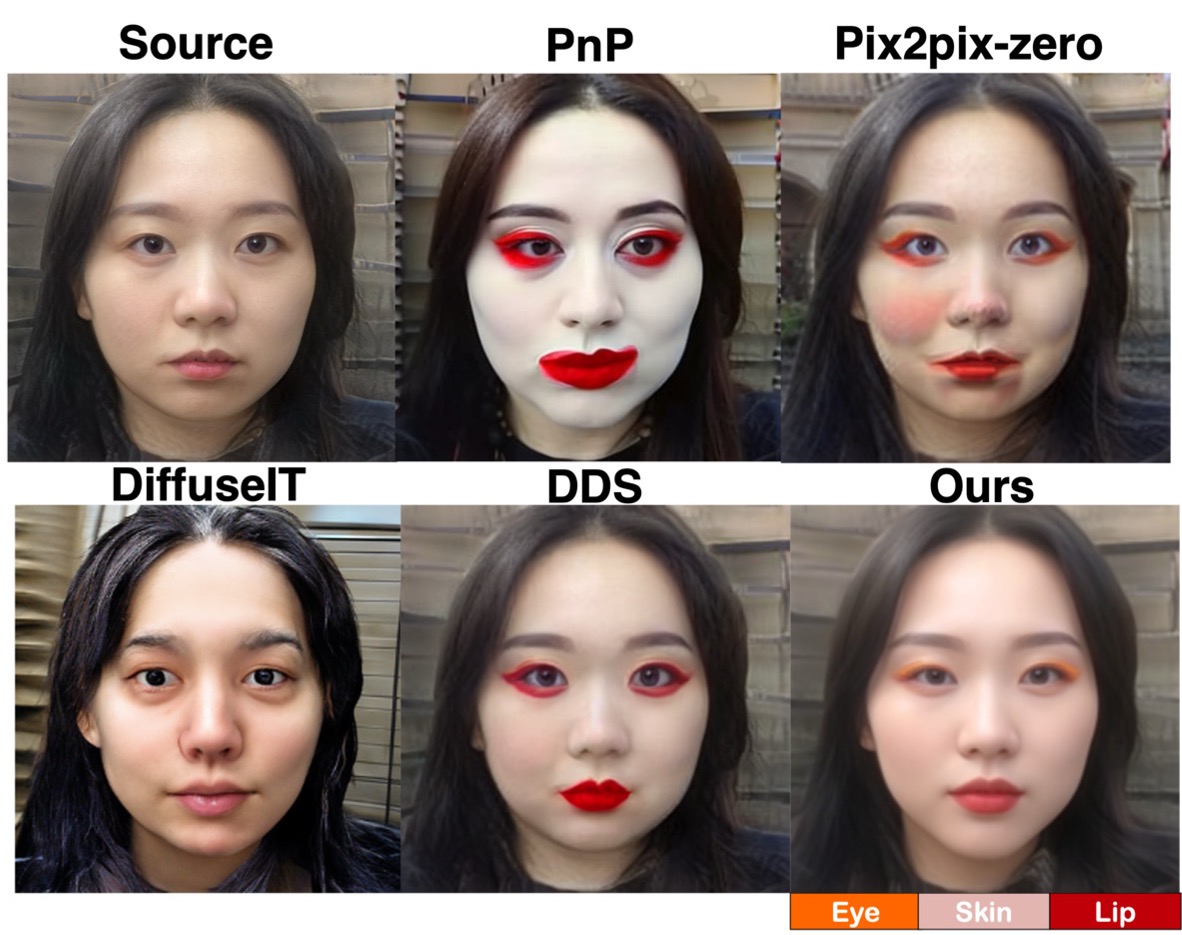}
    \caption{comparison with diffusion editing methods.}
    \vspace{-0.2cm}
    \label{fig:diffusion_editing}
\end{figure}

% \textcolor{blue}{
% \subsection{Comparison to diffusion-based editing methods}
% Regarding the comparison with the inversion-based framework, the proposed framework is notably efficient from various perspectives: inference speed, data requirements, training time, etc. Specifically, we early-stop DDIM inversion at $t^* = 200 < T$, in contrast to SOTA inversion-based frameworks (e.g., PnP, Pix2Pix-zero, EBCA, etc) which typically terminate at $t^* = T = 1000$. Additionally, our approach leverages \textit{one-step} pixel-domain guidance. 
% To quantitatively evaluate the efficiency, we compare the inference time and performance of text-based makeup customization against SOTA frameworks (note that existing frameworks are not opted for makeup customization). 
% \cref{tab:computational_cost} and \cref{fig:main}b highlight the advantageous computational costs and performance of our approach. 
% }

\begin{figure}[t!]
    \centering
    \includegraphics[width=\linewidth]{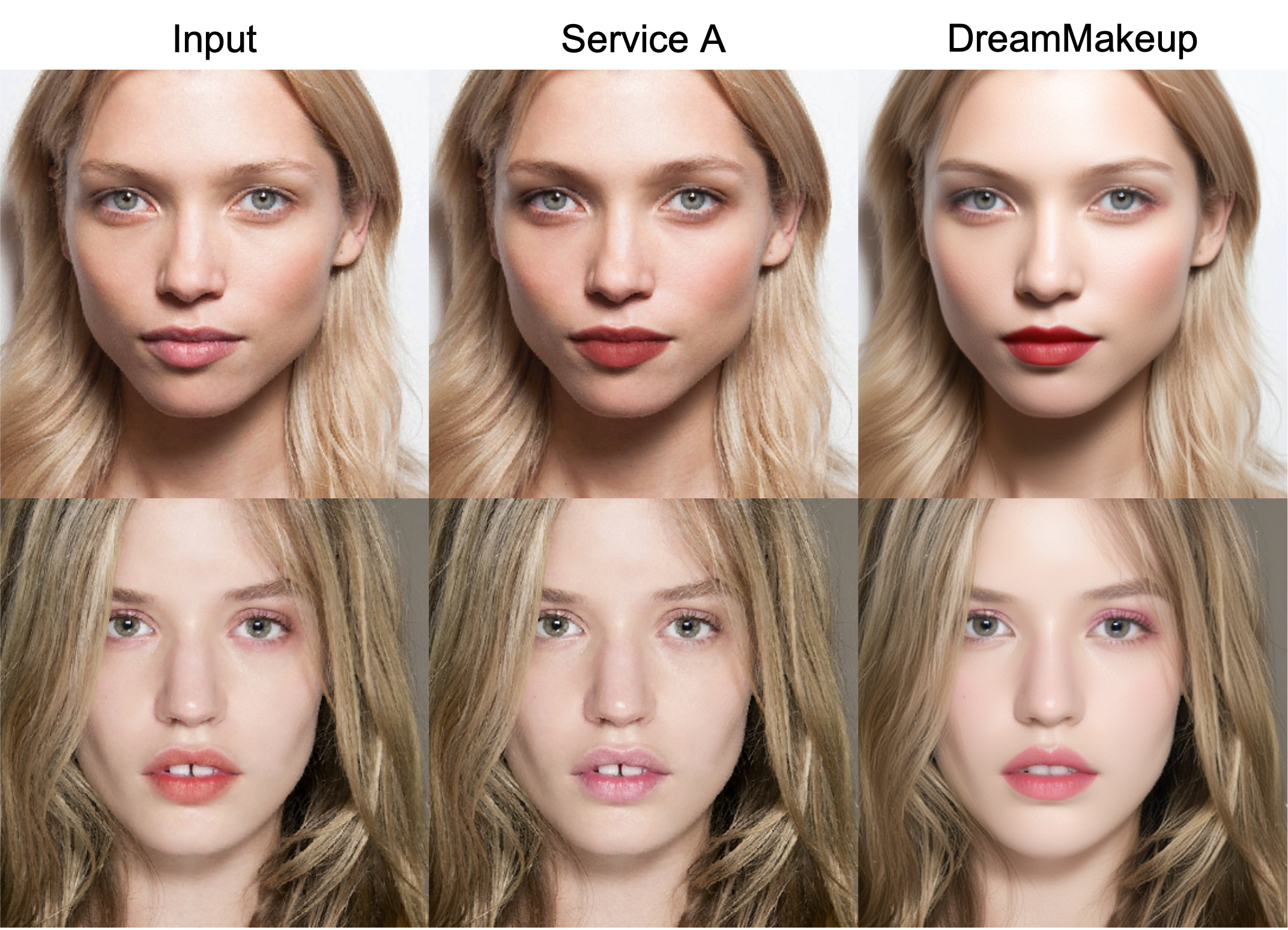}
    \caption{Comparisons of DreamMakeup with other global mobile AI makeup services.}
    \vspace{-0.6cm}
    \label{fig:meitu}
\end{figure}

\begin{figure}[tb]
    \centering
    \includegraphics[width=0.9\columnwidth]{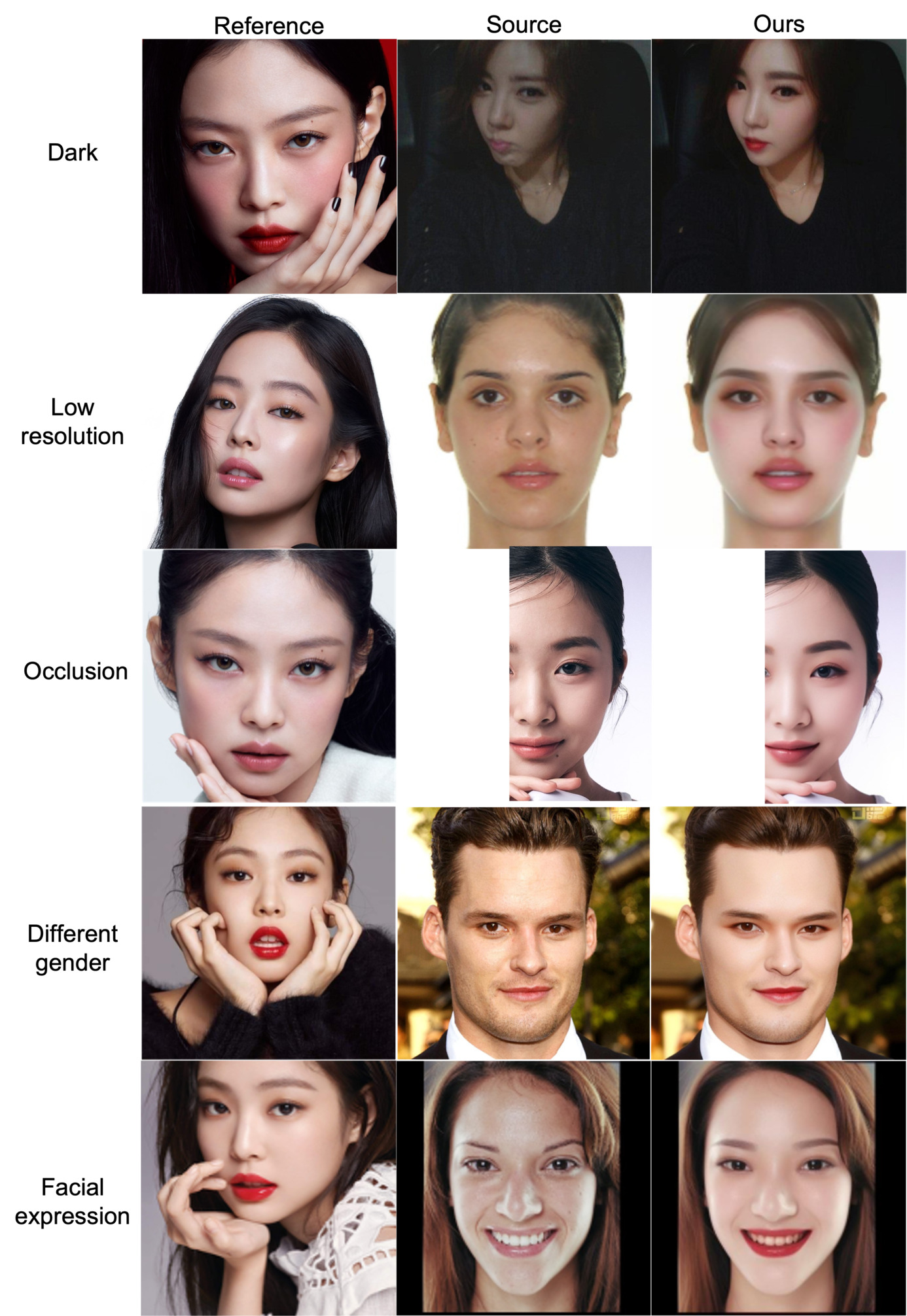}
    \caption{DreamMakeup results on extreme conditions.}
    \vspace{-0.4cm}
    \label{fig:extreme}
\end{figure}

\begin{figure}[tb]
    \centering
    \includegraphics[width=\columnwidth]{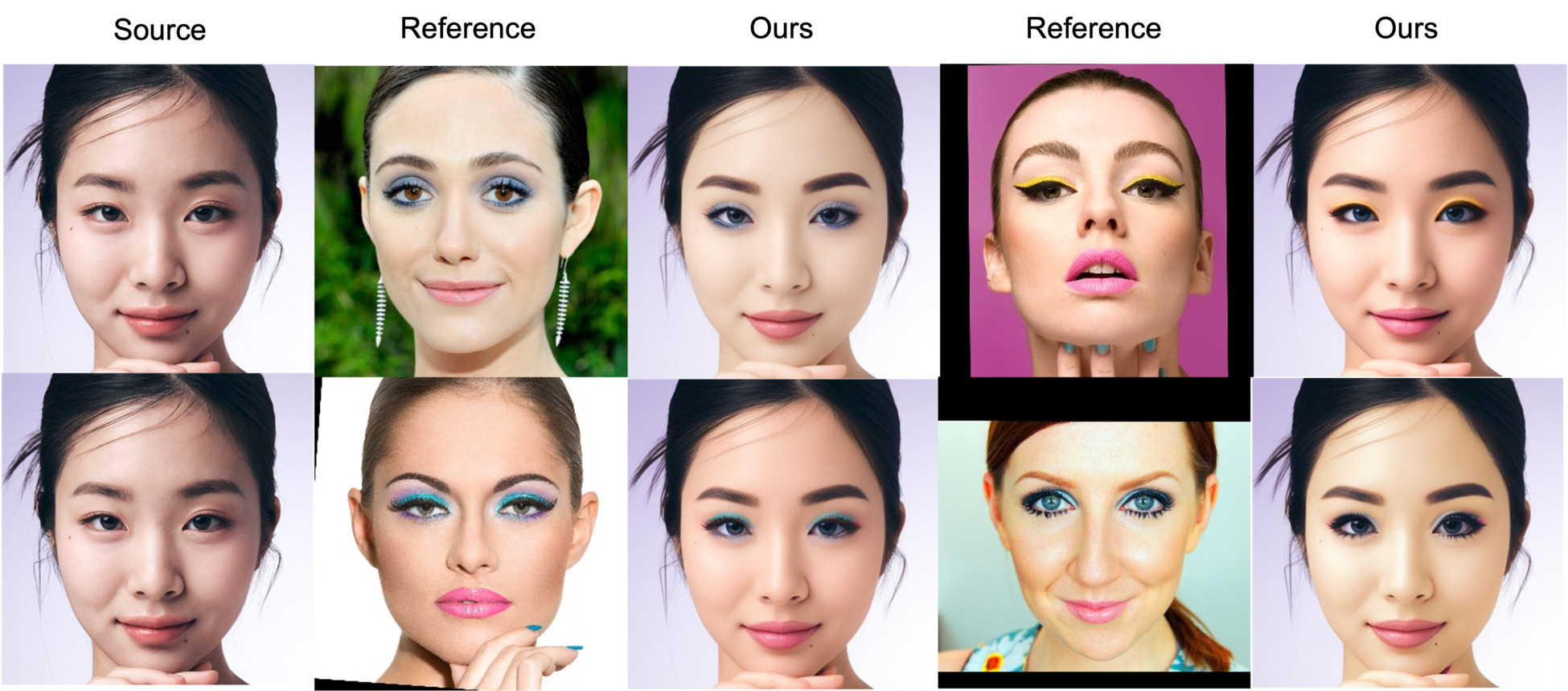}
    \caption{DreamMakeup results on extreme makeup. Reference images are from the LADN\cite{gu2019ladnlocaladversarialdisentangling} dataset.}
    \vspace{-0.4cm}
    \label{fig:extreme_makeup}
\end{figure}

\begin{figure}[tb]
    \centering
    \includegraphics[width=1.0\columnwidth]{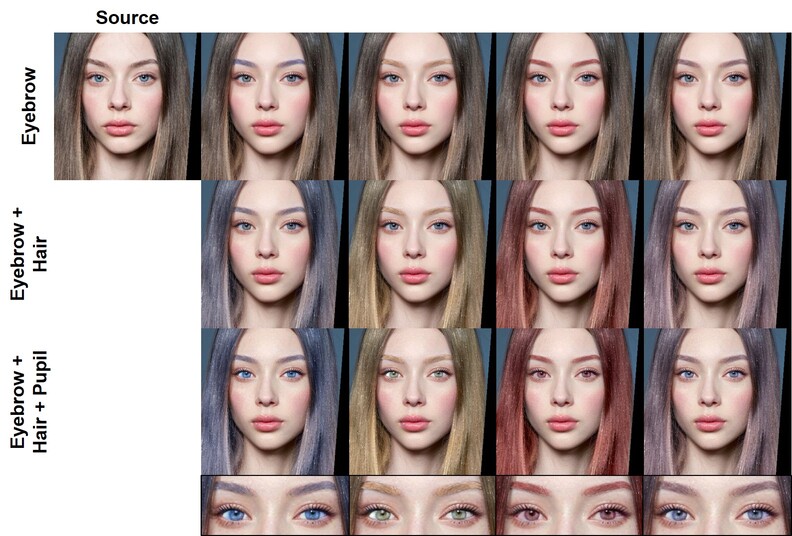}
    \caption{DreamMakeup results on RGB matching for eyebrows, pupils, and hair. We transform the color of each facial attribute within segmentation mask area.}
    \vspace{-0.4cm}
    \label{fig:RGB_application}
\end{figure}

\subsection{Additional Results}
To demonstrate the robustness of our method, we test its performance on a variety of challenging conditions, including images with low resolution, occlusions, and dark lighting, etc with results shown in \cref{fig:extreme}. Although our primary focus is on generating natural, daily makeup, the framework's flexibility allows it to be applied to more extreme and artistic makeup styles as well (\cref{fig:extreme_makeup}). Furthermore, the user can customize the target region for color transformation, extending the application of DreamMakeup beyond facial makeup to related tasks such as hair, eyebrow, and pupil coloring (\cref{fig:RGB_application}).

\section{Experimental details}
We use DDIM scheduler and set the early-stop inversion step ranging from $t^*=200$ to $t^*=400$. The number of reverse steps is set to $30$. LoRA scale $s$ is set to $0.2$. To smooth eye shadow masks, we employed a cross-shaped kernel with the size of $(12, 7)$ and performed $2$ iterations of mask dilation.
% In contrast, for eye masks, we conducted $2$ iterations of mask dilation using a rectangle-shaped kernel with a width of $15$.
% sampling step
The textual prompts commonly used in cross attention composition are as follows:
\begin{itemize}[noitemsep]
    \item natural lips, natural makeup, fair skin, asian skin
    \item korean makeup, korean style, korean beauty, (A Classy and Cute Korean girl:1.3), cute, (Korean idol), K-pop, skm\_misoo, beautiful
    \item 32K, high-res, (masterpiece:1.3), best quality, 8K.HDR, smooth face, 1 girl,close up face, (photorealistic:1.6), [:(detailed face:1.2):0.3]
    \item (Glossy lips:1.6), Gleaming lips, (fair skin:1.4), sharp focus, blusher
    \item (Goddess smile:1.3)
    \item (worst\_quality:2.0) low quality, blur, deformed ugly, pixelated, cgi, illustration, cartoon, deformed, distorted, disfigured, poorly drawn
\end{itemize}
The directional degree of $s$-th composition, $\alpha_s \le 0$, is assigned $0.1, 0.1, 0.3, 0.7, 0.1, -0.1$ for each prompt. 

\subsection{LLM}
To train the language model, we constructed a QnA dataset containing information matching makeup and facial attributes. The dataset consists of $460,000$ pairs of questions and answers. Below is an example of the makeup dataset.
%\vspace{0.3cm}

\noindent\texttt{\textcolor{darkgray}{\#\#\# Instruction: Which lip colors are suitable for women with the following condition? \textbackslash nbronze skin, square face, angular jaw}} \\
\noindent\texttt{\textcolor{darkgray}{\#\#\# Response: deep red or vivid red or dark red.} }
%\vspace{0.3cm}

As a base model, we utilized dolly-v2-3b\footnote{Databricks, Free dolly: Introducing the world’s first truly open instruction-tuned llm, https://github.com/databrickslabs/dolly, 2023.} and fine-tuned the model for 3 epochs using the makeup dataset. To prevent the model from forgetting language proficiency during fine-tuning, we also incorporated the natural language dataset used to train this base language model. The training objective is to generate the subsequent tokens based on the tokenized instructions in an autoregressive manner.

\section{Limitations}
While our proposed method offers an efficient, training-free framework for face makeup application via early-stop DDIM inversion, it requires multiple sampling timesteps to generate the final output. Since our method utilizes BiSeNet~\cite{yu2018bisenet} for facial segmentation and a pre-trained diffusion model for the generative process, our approach may inherit the intrinsic limitations of these foundational models.

\clearpage

%\begin{figure*}[t!]
%    \centering
%    \includegraphics[width=\textwidth]{Figure/rgb_comparison2.jpg}
%    \caption{The skin, lip, eye shadow makeup, and their combination.}
%    \label{fig:supp rgb_comparison}
%\end{figure*}

\end{document}

%% file: preamble.tex
\usepackage[utf8]{inputenc} % allow utf-8 input
\usepackage[T1]{fontenc}    % use 8-bit T1 fonts
\usepackage{url}            % simple URL typesetting
\usepackage{booktabs}       % professional-quality tables
\usepackage{amsfonts}       % blackboard math symbols
\usepackage{nicefrac}       % compact symbols for 1/2, etc.
\usepackage{microtype}      % microtypography
\usepackage{xcolor}       % colors
\usepackage{xkcdcolors}
\usepackage{times}
\usepackage{epsfig}
\usepackage{graphicx}
\usepackage{placeins}
\usepackage{multirow}
\usepackage[skip=5pt]{caption}
\usepackage{rotating}
\usepackage{cancel}
\usepackage{setspace}

\usepackage{bm}
\usepackage{bibunits} % for separate ref list in appendix

\usepackage{enumitem}
\usepackage{amssymb,amsthm}
\usepackage{amsmath}
\usepackage{graphicx}
\usepackage{wrapfig,lipsum,booktabs}

\usepackage{epsfig}
\usepackage{tikz}
\usetikzlibrary{spy}
\usepackage{algpseudocode}
\usepackage{algorithm}
\usepackage{mathrsfs}

\usepackage{nicefrac}       % compact symbols for 1/2, etc.
\usepackage{booktabs}       % professional-quality tables

\usepackage{thmtools,thm-restate}
\usepackage{listings}
\usepackage{lstautogobble}  %
\usepackage{color}          %
\usepackage{zi4}            %
\definecolor{bluekeywords}{rgb}{0.13, 0.13, 1}
\definecolor{greencomments}{rgb}{0, 0.5, 0}
\definecolor{redstrings}{rgb}{0.9, 0, 0}
\definecolor{graynumbers}{rgb}{0.5, 0.5, 0.5}
\usepackage{subcaption}        % subfigure
\usepackage{siunitx}               % <--- new
\usepackage[export]{adjustbox}
\usepackage{makecell}
\usepackage{url}
\usepackage{tikz, pgfplots}
\usetikzlibrary{positioning}
\usepackage{capt-of}
\usepackage{diagbox}

\def\eqref#1{(\ref{#1})}

\usepackage{colortbl}

\usepackage{mdframed}
\usepackage{transparent}

%% file: math_commands.tex
\usepackage{amsmath,amsfonts,bm}

\def\eqref#1{(\ref{#1})}
\def\1{\bm{1}}

\DeclareMathAlphabet{\mathsfit}{\encodingdefault}{\sfdefault}{m}{sl}
\SetMathAlphabet{\mathsfit}{bold}{\encodingdefault}{\sfdefault}{bx}{n}

\newcommand{\softmax}{\mathrm{softmax}}

\newcommand{\xb}{{\boldsymbol x}}

\newcommand{\zb}{{\boldsymbol z}}

\newcommand{\epsilonb}{{\boldsymbol \epsilon}}

\newcommand{\Dc}{{\mathcal D}}
\newcommand{\Ec}{{\mathcal E}}

\newcommand{\Nc}{{\mathcal N}}

\newcommand{\Tc}{{\mathcal T}}

\DeclareMathOperator{\data}{data}

\newcommand\norm[1]{\left\lVert#1\right\rVert}

\newcommand{\Cb}{{\boldsymbol C}}

\newcommand{\Rb}{{\mathbb R}}

\newcommand{\Eb}{{\mathbb E}}

\newcommand{\alphabar}{{\bar \alpha}}

\usepackage{amsthm}

\newcommand{\Ib}{{\boldsymbol I}}

\newcommand{\Q}{{\boldsymbol Q}}
\newcommand{\K}{{\boldsymbol K}}
\newcommand{\V}{{\boldsymbol V}}
\newcommand{\W}{{\boldsymbol W}}

\newcommand{\C}{{\boldsymbol C}}

\usetikzlibrary{positioning}
\usepackage{capt-of}
\usepackage{diagbox}

\definecolor{C0}{rgb}{0.121569, 0.466667, 0.705882}
\definecolor{C1}{rgb}{1.000000, 0.498039, 0.054902}
\definecolor{C2}{rgb}{0.172549, 0.627451, 0.172549}
\definecolor{C3}{rgb}{0.839216, 0.152941, 0.156863}
\definecolor{C4}{rgb}{0.580392, 0.403922, 0.741176}
\definecolor{C5}{rgb}{0.549020, 0.337255, 0.294118}
\definecolor{C6}{rgb}{0.890196, 0.466667, 0.760784}
\definecolor{C7}{rgb}{0.498039, 0.498039, 0.498039}
\definecolor{C8}{rgb}{0.737255, 0.741176, 0.133333}
\definecolor{C9}{rgb}{0.090196, 0.745098, 0.811765}
\definecolor{trolleygrey}{rgb}{0.5, 0.5, 0.5}

\definecolor{BrickRed}{rgb}{0.6,0,0}
\definecolor{RoyalBlue}{rgb}{0,0,0.8}
\definecolor{Tdgreen}{rgb}{0,0.4,0.7}
\definecolor{pinegreen}{rgb}{0.0, 0.47, 0.44}
\definecolor{cornellred}{rgb}{0.7, 0.11, 0.11}
\definecolor{cadmiumgreen}{rgb}{0.0, 0.42, 0.24}
\definecolor{spirodiscoball}{rgb}{0.06, 0.75, 0.99}
\definecolor{mylightblue}{rgb}{0.85, 0.90, 0.94}
\definecolor{maroon}{cmyk}{0,0.87,0.68,0.32}

\usepackage{pifont}% http://ctan.org/pkg/pifont
\algnewcommand{\LineComment}[1]{\State \(\triangleright\) #1}

\definecolor{cfg}{rgb}{0.906, 0.435, 0.318}
\definecolor{cfgpp}{rgb}{0.165, 0.616, 0.561}
\definecolor{cfgnull}{rgb}{0.208, 0.565, 0.953}

\definecolor{pp}{rgb}{0.165, 0.616, 0.561}
\definecolor{teacher}{rgb}{0.906, 0.435, 0.318}
\definecolor{student}{rgb}{0.208, 0.565, 0.953}
\definecolor{ppgreen}{rgb}{0.93, 0.98, 0.9}